\documentclass[sigconf]{acmart}
\makeatletter
\@ACM@balancefalse
\makeatother
\AtBeginDocument{%
  }

\usepackage[most]{tcolorbox}
\usepackage{multirow}
\usepackage{fancyvrb}
\usepackage{fvextra}
\usepackage[table]{xcolor}
\tcbset{
  promptbox/.style={
    boxrule=0.8pt,
    arc=2mm
  }
}

\copyrightyear{2026}
\acmYear{2026}
\setcopyright{cc}
\setcctype{by}

\acmConference[MM '26] {Proceedings of the 34th ACM International Conference on Multimedia}{November 10--14, 2026}{Rio de Janeiro, Brazil.}
\acmBooktitle{Proceedings of the 34th ACM International Conference on Multimedia (MM '26), November 10--14, 2026, Rio de Janeiro, Brazil}
\acmISBN{979-8-4007-2213-4/2026/11}
\acmDOI{10.1145/3767308.3836232}

\begin{document}

\title{When Helpful Text Hurts: Option-Redirecting Bias in Vision–Language Models}


\author{Tam Le Thi Thanh}
\orcid{0009-0009-5940-9656}
\email{23521386@gm.uit.edu.vn}
\affiliation{%
  \institution{University of Information Technology, VNU-HCM}
  \city{Ho Chi Minh City}
  \country{Vietnam}
}
\affiliation{%
  \institution{Vietnam National University}
  \city{Ho Chi Minh City}
  \country{Vietnam}
}

\author{Hoang Tran Van}
\orcid{0009-0007-9029-4442}
\email{23520542@gm.uit.edu.vn}
\affiliation{%
  \institution{University of Information Technology, VNU-HCM}
  \city{Ho Chi Minh City}
  \country{Vietnam}
}
\affiliation{%
  \institution{Vietnam National University}
  \city{Ho Chi Minh City}
  \country{Vietnam}
}

\author{Hong-Hanh Nguyen-Le}
\orcid{0000-0003-1553-2264}
\email{hong-hanh.nguyen-le@ucdconnect.ie}
\affiliation{%
  \institution{University College Dublin}
  \city{Dublin}
  \country{Ireland}
}

\author{Thanh Duc Ngo}
\authornote{Corresponding author.}
\orcid{0000-0001-6882-0070}
\email{thanhnd@uit.edu.vn}
\affiliation{%
  \institution{University of Information Technology, VNU-HCM}
  \city{Ho Chi Minh City}
  \country{Vietnam}
}
\affiliation{%
  \institution{Vietnam National University}
  \city{Ho Chi Minh City}
  \country{Vietnam}
}

\renewcommand{\shortauthors}{Tam Le Thi Thanh, Hoang Tran Van, Hong-Hanh Nguyen-Le, and Thanh Duc Ngo}

\begin{abstract}
   In tri-modal visual question answering (VQA), auxiliary text is commonly used to complement visual and textual inputs, yet its reliability is often uncontrolled. While prior work studies modality conflicts in general, it remains unclear how different types of unreliable auxiliary text affect answer selection under fixed image–question–option contexts. In this work, we show that the most harmful auxiliary text is not necessarily the most factually incorrect, but the one that aligns with the question while contradicting the image and favoring a specific distractor, leading to systematic redirection of model predictions. To isolate this effect, we introduce the Textual Reliability Ladder, a controlled diagnostic protocol that decomposes auxiliary text along three axes: image consistency, question relevance, and option support. Across multiple datasets (ScienceQA, VCR, A-OKVQA, Causal-VidQA) and recent VLMs, we find that such distractor-supporting text induces the largest accuracy drops (up to 53.1\%) and concentrates errors on specific incorrect options. To mitigate this failure mode, we propose a training-free inference-time intervention that explicitly counteracts this redirection effect via noise-stability steering and dynamic grounding, reducing redirected errors while largely preserving performance under faithful text. Our results highlight that auxiliary-text reliability must be understood at the decision level, rather than solely through factual correctness, and provide a practical pathway toward more robust tri-modal reasoning.
\end{abstract}

\ccsdesc[500]{Information systems~Question answering}
\keywords{Vision-Language Models, Visual Question Answering, Text Bias, Latent-space Steering}


\maketitle

\begin{figure}[!t]
   \centering
   \includegraphics[width=1\linewidth]{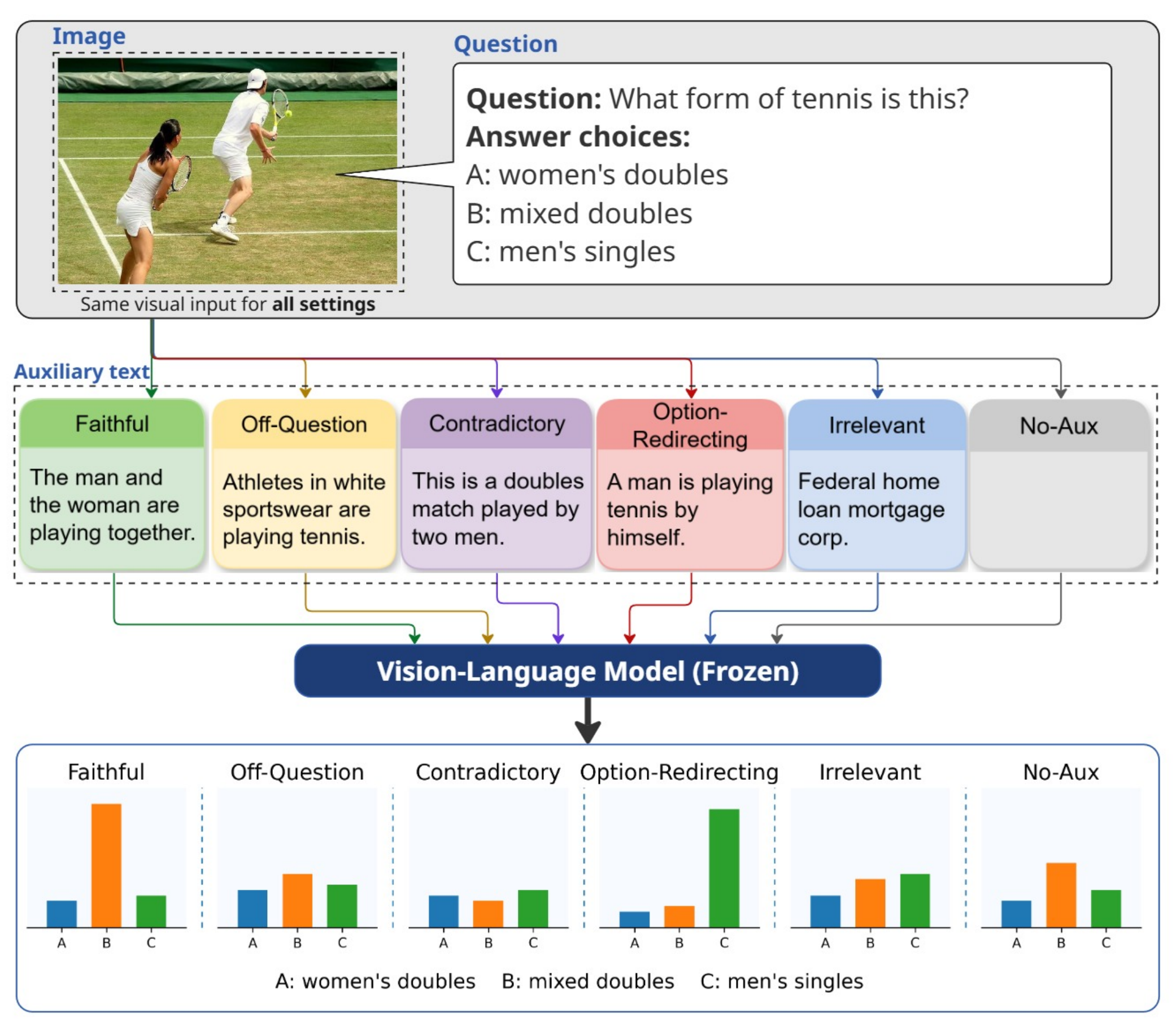}
   \caption{Decision-level auxiliary-text bias in tri-modal multiple-choice VQA. We fix the image, question, and answer options, and vary only the auxiliary text to isolate its effect on answer selection. While faithful text supports correct reasoning and irrelevant text induces mild uncertainty, certain misleading text can systematically redirect predictions toward a specific incorrect option. We categorize auxiliary text into five regimes—Faithful, Off-Question, Contradictory, Option-Redirecting, and Irrelevant—along with a No-Aux baseline. Among these, Option-Redirecting text is particularly harmful: although aligned with the question, it is inconsistent with the image and implicitly favors a distractor, leading to concentrated errors rather than diffuse confusion.}
   \Description{Schematic of tri-modal multiple-choice VQA. The image, question, and answer choices are fixed while the auxiliary text is varied across No-Aux, Faithful, Off-Question, Contradictory, Option-Redirecting, and Irrelevant conditions. Option-Redirecting text points toward a specific incorrect answer.}
   \label{fig:pipeline}
\end{figure}

\section{Introduction}

Recent visual question answering (VQA) systems increasingly operate in a tri-modal setting, where auxiliary text—such as captions, OCR outputs, or retrieved descriptions—is incorporated alongside images and questions to enhance reasoning. While such auxiliary signals can provide complementary context, their reliability is often uncontrolled in practice, particularly when generated or retrieved automatically. This raises a fundamental question: how does unreliable auxiliary text influence answer selection when the image, question, and candidate options are fixed, and can certain forms of such text systematically bias models toward specific incorrect answers?

Crucially, not all unreliable auxiliary text is equally harmful. While prior work \cite{deng2503words} treats textual noise or contradiction as a uniform source of error, we observe that the most damaging cases arise when auxiliary text is aligned with the question but inconsistent with the image, while supporting a specific incorrect option. In such cases, the model is not merely confused, but systematically redirected toward a particular distractor. This reveals a decision-level failure mode: auxiliary text can reshape the relative preference over candidate answers, concentrating errors on specific options rather than inducing diffuse uncertainty. As a result, evaluating auxiliary-text reliability solely in terms of factual correctness or modality agreement is insufficient to explain its impact on model behavior.

This behavior is closely tied to how auxiliary text is integrated into VLMs. Textual inputs encode high-level objects, attributes, and relations that are naturally aligned with the semantics of VQA tasks. When accurate, such signals act as informative priors that facilitate reasoning. However, when auxiliary text is incorrect but remains question-aligned, it can inject misleading evidence through the same fusion pathways, steering attention and scoring toward linguistically plausible yet visually unsupported answers. Consequently, the most harmful auxiliary text is not the most obviously incorrect, but the one that is semantically aligned in the wrong direction.

To systematically analyze this phenomenon, we introduce the \textbf{Textual Reliability Ladder}, a controlled diagnostic framework for studying auxiliary-text effects in tri-modal VQA. The framework varies only the auxiliary text while holding the image, question, and answer options fixed, enabling direct attribution of changes in model predictions to textual properties. Specifically, the ladder decomposes auxiliary text along three key dimensions—image consistency, question relevance, and option support—allowing the construction of distinct reliability conditions under a shared context. This design provides a structured way to isolate and compare the effects of different types of auxiliary text on answer selection.

Using this framework, we identify Option-Redirecting auxiliary text as a particularly harmful and under-explored failure mode across multiple VQA benchmarks and frozen VLMs. Such text induces the largest performance degradation and systematically concentrates errors on a specific incorrect option, even under style control and lexical filtering. In contrast, other unreliable conditions—such as off-question or weakly grounded descriptions—tend to produce smaller and less consistent effects.

Motivated by these insights, we further explore a lightweight inference-time intervention designed to mitigate Option-Redirecting bias while preserving model behavior under reliable auxiliary text. The proposed approach operates via post-hoc latent-space steering and requires only access to intermediate activations, without gradients or parameter updates. We emphasize that this intervention is not intended as a universal solution, but as a targeted mitigation aligned with the specific failure mode revealed by our analysis.

Our contributions are threefold:
\begin{itemize}
    \item We introduce the Textual Reliability Ladder, a controlled diagnostic framework that isolates decision-level auxiliary-text bias in tri-modal multiple-choice VQA.
    \item We identify Option-Redirecting auxiliary text as a particularly damaging regime that induces systematic, option-targeted errors rather than diffuse uncertainty.
    \item We propose a lightweight inference-time method that mitigates such failures under well-defined access assumptions, while largely preserving performance under reliable auxiliary text conditions.
\end{itemize}

\section{Related Work}

\subsection{Bias from Auxiliary Text}

Surveys on VLMs hallucination typically define them as misalignment between visual content and model-generated text, summarizing causes, benchmarks, and mitigation strategies \cite{bai2024hallucination,liu2024survey}. In practical VQA pipelines, however, models often consume \emph{auxiliary text provided as input} by upstream components or retrieval/augmentation components (e.g., captions, OCR/ASR, metadata) \cite{jian2024large,jiang2025memory,chen2025livecc}. Such text can omit key attributes or introduce plausible but spurious details, especially in text-rich visual understanding where OCR/layout cues strongly affect reasoning \cite{fu2024ocrbench,yu2025cross}. We therefore study \emph{text bias}, defined as decision-level over-reliance on auxiliary text in tri-modal VQA (image, question, auxiliary text), where unreliable textual inputs can distort discrete answer selection. Unlike prior work on hallucination, which evaluates the faithfulness of generated text, our focus is on how imperfect auxiliary inputs affect downstream decisions. 


Recent work studies \emph{modality conflict}—cases where image evidence and auxiliary text provide incompatible signals—to analyze modality preference and conflict resolution in VLMs \cite{hua2025vision,zhang2025robust,nguyen2025challenges,deng2503words,liu2024insight}. Conflict-focused benchmarks (e.g., CrossCheck) show that VLMs often struggle to detect and resolve contradictions, leading to text-dominant failures under misleading captions \cite{popordanoskacrosscheck,tian2025crosscheck,deng2503words,zhang2025robust}. However, these approaches primarily evaluate whether models can recognize \emph{contradiction}, rather than how auxiliary text influences \emph{which incorrect answer is selected}. As a result, they do not explicitly capture decision-level effects where errors are concentrated on specific distractors.


Similarly, prior work on language priors and shortcut behaviors in multiple-choice VQA shows that superficial textual cues can bias models toward particular answers, motivating debiasing methods that reduce unimodal reliance \cite{agrawal2018don,cadene2019rubi,guo2021loss,varma2024ravl,zhang2025mitigating}. Mechanistic analyses further suggest that visual evidence may remain internally accessible, yet the model may fail to exploit it at the decision stage, amplifying reliance on textual cues \cite{fu2025hidden}. Additional studies show that injected textual signals, such as captions or OCR, can further steer multimodal decisions \cite{tao2025imgtrojan,qraitem2025web}. Importantly, these approaches primarily focus on modality-level conflicts or dataset-level biases, and do not explicitly model instance-level effects induced by externally provided auxiliary text.


Despite these advances, prior diagnostics largely focus on contradiction detection or aggregate robustness under broad text-corruption settings \cite{popordanoskacrosscheck,tian2025crosscheck,deng2503words,zhang2025robust}. They do not isolate which properties of \emph{input-level} auxiliary text drive decision-level bias in tri-modal \emph{multiple-choice} VQA, nor do they distinguish between diffuse noise and \emph{option-targeted} redirection toward a specific distractor. 

To address this gap, we introduce controlled auxiliary-text variants that factorize (i) image grounding, (ii) question alignment, and (iii) option-targeted support, enabling analysis of how auxiliary text shapes discrete answer selection under a fixed context.

\subsection{Latent-Space Steering and Test-Time Intervention}

Latent-space steering modifies internal representations at inference time to control model behaviors without retraining \cite{huang2024opera}. In VLMs, it has mainly been used to reduce hallucinations or strengthen visual grounding by nudging hidden states toward visually consistent regions \cite{liu2025reducing,yang2025nullu}. Related works explore steering for hallucination detection \cite{park2025steer}, transferring steering vectors to improve multimodal understanding \cite{gan2025textual}, and activation-level control as a general inference-time mechanism \cite{sivakumar2025steervlm}. Complementary test-time methods adapt prompts online \cite{shu2022test,yoon2024c,sharifdeen2025tpt} or adjust decoding/calibration to strengthen visual grounding \cite{leng2024mitigating,zhu2025ibd}, while post-hoc debiasing of language priors has been studied in related scoring setups \cite{lin2023revisiting}. Unlike prior methods that aim to improve general grounding or reduce hallucination, our intervention specifically targets decision-level bias induced by auxiliary text.

Many existing interventions target hallucination-centric objectives or improve accuracy/calibration under trusted prompting \cite{liu2025reducing,yang2025nullu,shu2022test}. In contrast, we focus on \emph{decision-level} text bias in tri-modal \emph{multiple-choice} VQA, where auxiliary text can be question-relevant yet image-inconsistent and selectively support a particular answer option. Accordingly, we analyze how auxiliary text \emph{shifts discrete predictions} and increases the likelihood of selecting text-supported (potentially incorrect) options under controlled conditions. 
As fine-tuning confounds ladder attribution by entangling text-reliability effects with parameter updates (as explained in Appendix~\ref{app:why_no_finetune}), we restrict our study to frozen VLMs and design a test-time intervention.

\begin{table}[t]
\centering
\setlength{\tabcolsep}{2pt}
\renewcommand{\arraystretch}{1.15} 
\begin{tabular}{lccccl}
\toprule
\rowcolor{gray!10}
Textual type & Image & Question & Wrong ans. & Correct ans. \\
\midrule
Faithful         & $\checkmark$ & $\checkmark$ &        $\times$  & $\checkmark$ \\
Off-Question   & $\checkmark$ & $\times$ & $\times$ &       $\times$ \\
Contradictory  & $\times$     & $\checkmark$ & $\times$ &    $\times$  \\
Option-Redirecting        & $\times$     & $\checkmark$ & $\checkmark$ &     $\times$  \\
Irrelevant       & $\times$     & $\times$     &        $\times$  & $\times$ \\
\bottomrule
\end{tabular}%
\caption{Textual Reliability Ladder conditions and their diagnostic cues. Each condition is defined by three factors: whether the auxiliary text is grounded in the \emph{image}, relevant to the \emph{question}, and whether it implicitly supports a specific \emph{correct} or \emph{incorrect} answer option. The ladder spans from reliable (Faithful), to non-informative (Off-Question, Irrelevant), to decision-level misleading (Option-Redirecting).}

\label{tab:textual-conditions}
\end{table}

\section{Textual Reliability Ladder}
\label{sec:ladder}

To systematically investigate how VLMs balance visual evidence with auxiliary text, we introduce the
\textbf{Textual Reliability Ladder}, a controlled diagnostic protocol that varies the reliability of auxiliary
text while keeping the \emph{image, question, and answer options} fixed.
This controlled setup enables direct attribution of prediction changes to the auxiliary text alone.

We focus on multiple-choice VQA, where discrete options enable decision-level analysis: auxiliary text is considered helpful when it increases the likelihood of the correct option, and harmful if it redirects probability mass toward a distractor. For each image--question pair, we construct auxiliary texts with controlled relationships to the image, the question, and the candidate answers, enabling fair comparisons where behavioral changes are attributable to textual properties rather than visual or question variation.

As shown in Table~\ref{tab:textual-conditions}, the ladder spans from fully reliable to misleading auxiliary text by organizing three diagnostic cues: visual grounding, question relevance, and option-targeted support. The ladder is not intended as an exhaustive $2^3$ enumeration of all binary cue combinations. Rather, it retains only semantically coherent and diagnostically useful regimes for multiple-choice VQA. These regimes are selected to isolate qualitatively distinct and practically relevant failure modes observed in VLM behavior, rather than to exhaust all combinatorial possibilities. \textit{No-Aux} is reported separately as a text-free baseline. 

We exclude the case where auxiliary text is both image-grounded and question-aligned yet supports neither the correct nor any incorrect option, because such text typically carries at least some answer-relevant evidence; forcing a fully support-neutral version would be weakly specified and often collapse into a borderline \textit{Faithful} or \textit{Off-Question} instance. We also exclude all cases in which text supports an answer option without being aligned to the current question, whether or not it is image-grounded. Option-level support is defined with respect to the question; without question alignment, apparent support reduces to a superficial option cue or leakage signal rather than a meaningful auxiliary-text regime. Thus, the ladder should be read as a controlled diagnostic subset of coherent regimes, rather than a full combinatorial product.

\begin{itemize}
    \item \textbf{Faithful}: grounded in the image and aligned with the question; provides evidence that supports the correct option and serves as a reliable reference condition for assessing the effect of auxiliary text.
    
    \item \textbf{Off-Question}: grounded in the image but irrelevant to the question; provides visually plausible yet task-irrelevant information and is not expected to systematically favor any answer option, primarily introducing distraction rather than targeted bias.
    
    \item \textbf{Contradictory}: aligned with the question but inconsistent with the image; induces cross-modal conflict \emph{without providing option-targeted support}, and is intended to produce non-targeted confusion rather than redirection toward any specific option.
    
    \item \textbf{Option-Redirecting}: aligned with the question but inconsistent with the image, and constructed to provide \textit{implicit, descriptive evidence consistent with a specific incorrect answer option}, without directly revealing or matching the option text, thereby inducing redirection of probability mass toward a single distractor.
    
    \item \textbf{Irrelevant}: unrelated to both the image and the question; serves as a pure noise baseline that provides no meaningful evidence for any answer option and is not expected to induce systematic decision bias.
    
    \item \textbf{No-Aux}: no auxiliary text is provided; serves as a text-free baseline where predictions rely solely on the image and the question.
\end{itemize}

To implement the ladder while minimizing confounds, we use a two-step hybrid pipeline: (i) constrained generation, and (ii) human verification in Appendix~\ref{app:human_verification}. These controls are designed to isolate the effect of semantic properties of auxiliary text, rather than superficial artifacts. Specifically, we target three common failure modes in multiple-choice VQA: \emph{style or length artifacts}, \emph{shallow option-string cues}, and \emph{answer leakage}.

\textbf{Faithful} texts are sourced from dataset-provided annotations (see Appendix~\ref{app:Faithful}). Rather than regenerating text when reliable textual reasoning is already available, we reuse these annotations and apply strict per-instance screening to verify that the retained Faithful texts satisfy our ladder constraints. This avoids an unnecessary generation stage, reduces additional cost and variance, and keeps the Faithful condition close to naturally occurring auxiliary supervision. If a Faithful annotation triggers the option-cue or leakage filters (e.g., it contains an answer string or strongly matches an option), we remove the entire image--question instance from the ladder, rather than editing the text, to preserve strict comparability across all auxiliary-text variants and avoid introducing asymmetric preprocessing.

\textbf{Unreliable} texts (\textbf{Off-Question} / \textbf{Contradictory} / \textbf{Option-Redirecting}) are generated using Gemini-2.5-Flash \cite{comanici2025gemini} with fixed prompt templates (see Appendix~\ref{app:off_question}, \ref{app:Contradictory}, \ref{app:option_redirecting}).  Option-Redirecting is token-length matched to the corresponding Faithful text on a per-instance basis to reduce the influence of superficial length effects in the key comparison. Other conditions are generated under fixed prompts, human verification, and lexical/style controls, with only coarse length bounds rather than exact matching.

For \textbf{Option-Redirecting}, we randomly select a target distractor $o_{\text{redir}}$ from the incorrect options and instruct the generator to produce \textbf{descriptive statements that are semantically consistent with $o_{\text{redir}}$} while contradicting the image. To avoid trivial string matching or explicit answer leakage, we require that redirected texts do \emph{not} mention the target option verbatim and instead provide implicit, descriptive support (e.g., attributes or relations consistent with $o_{\text{redir}}$) that may increase its likelihood at the decision stage without explicitly encoding the answer.

\textbf{Irrelevant} texts are sampled from WikiText passages \cite{merity2016pointer} and filtered to avoid accidental overlap with the question or options (lexical-similarity, option-cue, and leakage filters) in Appendix~\ref{app:irrelevant}. We also match the instance-wise style window used in other ladder conditions. This yields auxiliary text that is effectively uninformative for both the question and the decision while controlling for style confounds.

\paragraph{Human verification.}
All constructed texts undergo mandatory human verification using a checklist explicitly aligned with the ladder definition:
(i) image consistency versus contradiction as required,
(ii) question alignment versus off-question behavior,
(iii) absence of option-targeted support when not intended,
(iv) presence of option-targeted support for \textbf{Option-Redirecting} without verbatim option leakage, and
(v) absence of answer leakage and invalid cues.
Texts that are ambiguous or non-compliant with these criteria are removed and regenerated. We ensure consistency through multi-annotator agreement and iterative validation (see Appendix~\ref{app:human_verification}).

In contrast to common practice that groups diverse textual corruptions into a single undifferentiated category,
the Textual Reliability Ladder explicitly separates distinct failure mechanisms—including off-question distraction,
non-targeted contradiction, and option-targeted misdirection—under a controlled, multiple-choice setting (an illustrative example is shown in Appendix~\ref{app:example_ladder}). This factorized design enables decision-level diagnosis of how auxiliary text influences discrete predictions, rather than conflating qualitatively different behaviors into aggregate robustness scores. Finally, we evaluate frozen VLMs to avoid training-time confounds: behavioral changes observed under the ladder are primarily attributable to the influence of auxiliary text, rather than altered supervision exposure or parameter updates.

\section{Inference-Time Intervention}

\begin{figure*}[t!]
    \centering
    \includegraphics[width=\linewidth]{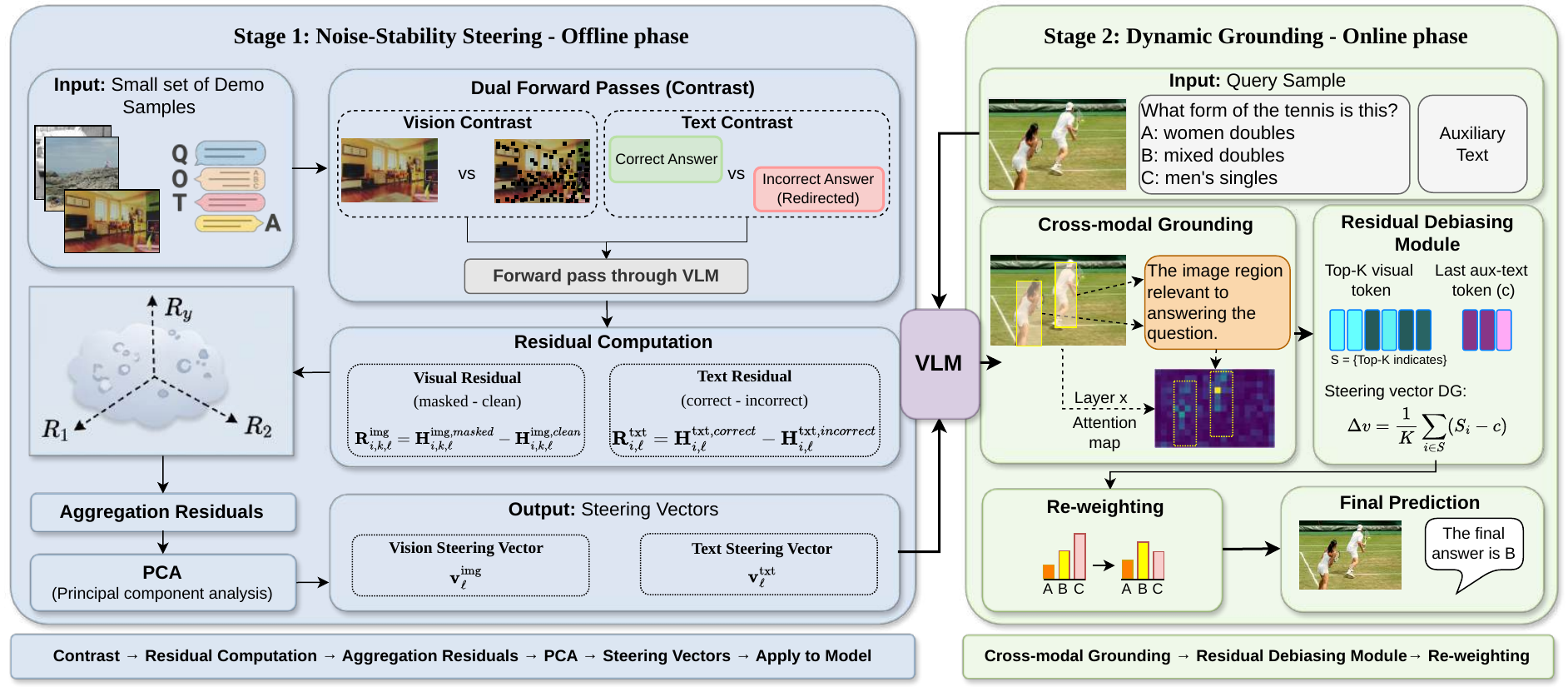}
    \caption{Overview of our two-stage inference-time intervention. Stage 1 estimates global steering directions from contrastive residuals, while Stage 2 performs instance-level, question-aware grounding by aligning textual representations with question-relevant visual evidence. Together, the two stages reduce option-redirecting auxiliary-text bias while maintaining performance under non-harmful auxiliary text.}
    
    \Description{Flow diagram of the two-stage inference-time intervention. Stage 1 computes global Noise-Stability Steering directions offline from contrastive residuals, and Stage 2 performs online Dynamic Grounding by aligning auxiliary-text representations with question-relevant visual evidence before answer prediction.}
    
    \label{fig:pipeline_method}
\end{figure*}

Auxiliary text can bias VLM decisions in two distinct ways: (i) \emph{global text dominance}, where the model becomes overly sensitive to textual input across instances, and (ii) \emph{instance-specific grounding conflicts}, where question-relevant visual evidence is overridden by misleading auxiliary text. Under reliable auxiliary text, the model is expected to assign high probability to the correct option. However, under \textbf{Option-Redirecting} text, probability mass is often shifted toward the corresponding distractor option instead.

Our goal is to mitigate such text-induced option bias without modifying model parameters or retraining the model. To address both global and instance-level effects in frozen VLMs, we propose a two-stage coarse-to-fine inference-time intervention, illustrated in Figure~\ref{fig:pipeline_method}.

We deliberately address misleading auxiliary text via \emph{steering} rather than regeneration or rewriting. If misleading text were first repaired upstream, the problem would shift from evaluating model robustness under diverse auxiliary-text regimes to a text-repair pipeline, making it harder to isolate the decision-level failure mode revealed by the ladder. By keeping the auxiliary text fixed and intervening only at inference time, we are able to examine whether a frozen VLM can resist misleading textual evidence, rather than benefiting from a sanitized replacement.

\begin{itemize}
    \item \textbf{Stage 1 (Noise-Stability Steering, offline):} a global, static representation shift estimated from paired correct and incorrect demonstrations, applied uniformly across instances.
    
    \item \textbf{Stage 2 (Dynamic Grounding, online):} an instance-level, question-aware intervention that adaptively realigns text representations with visual evidence.
\end{itemize}

\subsection{Stage 1: Noise-Stability Steering (NSS) with Offline Pre-computation and Pre-test Incorporation}


NSS leverages representation differences induced by correct versus incorrect answer continuations under a fixed (image, question, options) context. The underlying assumption is that correctness-related variation constitutes a consistent and dominant factor across demonstrations, allowing these differences to capture a shared direction associated with reliable versus misleading textual influence. Modality-specific steering vectors are estimated offline from a small dataset-specific demonstration pool and applied to the frozen VLM via static forward hooks, without parameter updates.


For each benchmark, we sample $N_{\text{demo}}$ demonstrations from the same dataset (e.g., a training split or held-out pool), ensuring that they are disjoint from evaluation instances. The value of $N_{\text{demo}}$ is selected to provide a stable estimate of representation differences while maintaining reasonable computational cost.

Each demonstration is a tri-modal instance paired with two answer continuations: $d_i=(I_i, Q_i, \mathcal{O}_i, T_i, A_i^+, A_i^-)$, where $I_i$ are the input image(s), $Q_i$ is the question, $\mathcal{O}_i=\{o_i^{(1)},\dots,o_i^{(M)}\}$ is the set of multiple-choice options, and $T_i$ is the \textit{Option-Redirecting} text targeting a distractor index $m^- \neq m^+$.
Both $A_i^+$ and $A_i^-$ are verbalized with the same fixed template, \emph{``The correct option is (\underline{m}): }\texttt{<option-text>}\emph{''}: $A_i^+$ uses the ground-truth index $m^+$ and option text $o_i^{(m^+)}$, while $A_i^-$ uses the redirected distractor index $m^-$ and option text $o_i^{(m^-)}$ from the same instance. An illustrative example is provided in Appendix~\ref{app:demo_samples}.


For each demonstration $d_i$, we estimate modality-specific steering directions via \emph{paired} forward passes that hold the full context fixed and change only one factor at a time.

\paragraph{\textbf{Vision (noise-effect residual}).} We run two passes with identical textual context (question, options, and auxiliary text), but different image tokens: a clean-image pass $(+)$ and a randomly masked-image-token pass $(-)$ (see Appendix~\ref{app:demo_hparams1} for full details). The masking operation perturbs visual evidence while preserving the textual context, allowing us to isolate representation changes associated with degraded visual grounding.

For multi-image inputs $I_i=\{I_{i,k}\}_{k=1}^{K_i}$, let $\mathbf{H}^{\text{img},+}_{i,k,\ell}$ and
$\mathbf{H}^{\text{img},-}_{i,k,\ell}$ be the post-MLP hidden states at layer $\ell$ restricted to the image tokens of image $k$. We define the visual residual

\begin{equation}
\mathbf{R}^{\text{img}}_{i,k,\ell}=\mathbf{H}^{\text{img},-}_{i,k,\ell}-\mathbf{H}^{\text{img},+}_{i,k,\ell},
\end{equation}
and aggregate to a single $d$-dimensional vector per instance/layer by mean-pooling over token positions and averaging across images:

\begin{equation}
\mathbf{r}^{\text{img}}_{i,\ell}
=\frac{1}{K_i}\sum_{k=1}^{K_i}\mathrm{MeanPool}\!\left(\mathbf{R}^{\text{img}}_{i,k,\ell}\right)\in\mathbb{R}^{d}
\end{equation}

\paragraph{\textbf{Text (correctness residual).}}
We run two passes that share the same image and prompt context, and differ \emph{only} in the answer continuation: a correct continuation $A_i^+$ versus a redirected incorrect continuation $A_i^-$. Let $\mathbf{H}^{\text{txt},+}_{i,\ell}$ and $\mathbf{H}^{\text{txt},-}_{i,\ell}$ denote the post-MLP hidden states at layer $\ell$ restricted to the aux-text token stream in the text decoder. The textual residual is
\begin{equation}
\mathbf{R}^{\text{txt}}_{i,\ell}=\mathbf{H}^{\text{txt},+}_{i,\ell}-\mathbf{H}^{\text{txt},-}_{i,\ell}\in\mathbb{R}^{n_{\text{txt}}\times d}
\end{equation}

We extract a single (rank-1) direction per layer and modality using PCA.
For \textbf{vision}, we apply PCA to the instance-level pooled residuals
$\{\mathbf{r}^{\text{img}}_{i,\ell}\}_{i=1}^{N_{\text{demo}}}$ and take the first principal component:
\begin{equation}
\mathbf{v}^{\text{img}}_{\ell}=\mathrm{PC}_1\!\left(\{\mathbf{r}^{\text{img}}_{i,\ell}\}_{i=1}^{N_{\text{demo}}}\right)\in\mathbb{R}^{d}
\end{equation}
For \textbf{text}, we compute PCA directly on token-level correctness residuals aggregated across demonstrations:
\begin{equation}
\mathcal{S}^{\text{txt}}_{\ell}=\bigcup_{i=1}^{N_{\text{demo}}}\{\mathbf{R}^{\text{txt}}_{i,\ell}[j,:]\}_{j=1}^{n_{\text{txt}}},   
\mathbf{v}^{\text{txt}}_{\ell}=\mathrm{PC}_1\!\left(\mathcal{S}^{\text{txt}}_{\ell}\right)\in\mathbb{R}^{d}
\end{equation}

We incorporate NSS into the frozen VLM via forward hooks \emph{once} prior to the test phase.
At all layers, we (i) improve noise-stability in the vision encoder by subtracting the visual direction and
(ii) encourage representations in the text decoder that are associated with correct continuations by adding the textual direction:
\begin{equation}
\mathbf{h}^{(\ell)\prime}_{\mathrm{post}}
=\mathbf{h}^{(\ell)}_{\mathrm{post}}
+
\begin{cases}
-\alpha_{\text{img}}\,\mathbf{v}^{\text{img}}_{\ell}, & \text{vision encoder}\\[3pt]
\ \alpha_{\text{txt}}\,\mathbf{v}^{\text{txt}}_{\ell}, & \text{text decoder}
\end{cases}
\end{equation}

Once enabled, NSS requires no additional demonstrations or extra forward passes at test time: cached directions are applied uniformly to each test sample. All NSS hyperparameters (e.g., PCA rank, normalization, scaling/clipping, and strengths $\alpha_{\text{img}},\alpha_{\text{txt}}$) are reported in Appendix~\ref{app:demo_hparams1}.
We compute and cache $\{\mathbf{v}^{\text{img}}_{\ell}\}$ and $\{\mathbf{v}^{\text{txt}}_{\ell}\}$ offline once.

\begin{figure*}[!t]
    \centering
    \includegraphics[width=\linewidth]{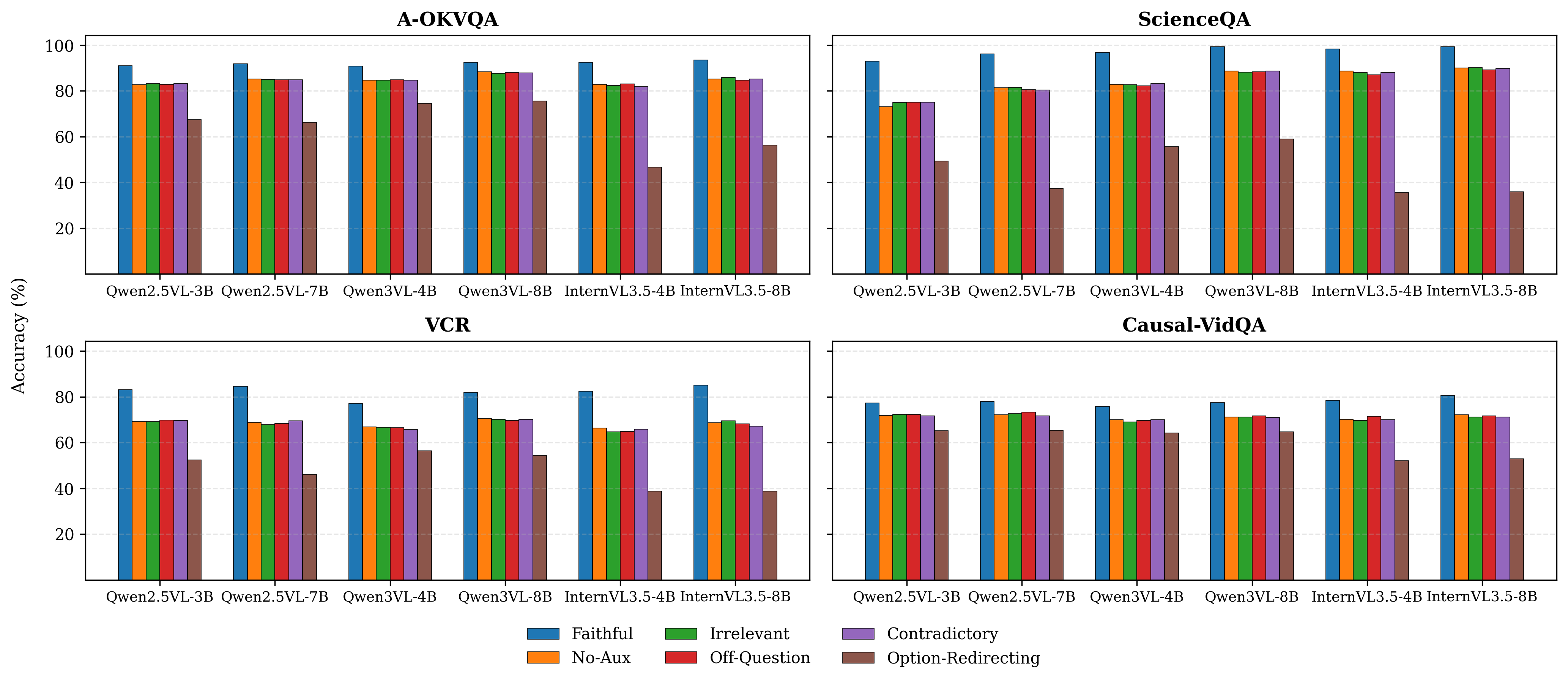}
        \caption{Accuracy of frozen VLMs under the Textual Reliability Ladder. Option-Redirecting text causes the largest and most
consistent degradation across datasets and models}
    \Description{Performance plots across A-OKVQA, ScienceQA, VCR, and Causal-VidQA for several frozen vision-language models under the Textual Reliability Ladder. Option-Redirecting text produces the largest and most consistent accuracy decrease.}
    \label{fig:textual_ladder}  
\end{figure*}

\subsection{Stage 2: Dynamic Grounding (DG) with Online Test-Time}

NSS provides a global, static correction, but it cannot address \emph{instance-specific} grounding conflicts where the
auxiliary text is question-relevant yet visually inconsistent. We therefore introduce DG,
a lightweight online intervention applied per test instance to suppress unreliable auxiliary-text influence.

At decoder layer $\ell$, let
$\mathbf{H}^{\mathrm{q}}_{\ell}\in\mathbb{R}^{T_q\times d}$,
$\mathbf{H}^{\mathrm{img}}_{\ell}\in\mathbb{R}^{T_{\mathrm{img}}\times d}$, and
$\mathbf{H}^{\mathrm{aux}}_{\ell}\in\mathbb{R}^{T_{\mathrm{aux}}\times d}$
denote post-MLP hidden states for question, image, and auxiliary-text tokens, respectively.
We also denote the hidden states of the \emph{text stream} (question + auxiliary text) by
$\mathbf{H}^{\mathrm{txt}}_{\ell}\in\mathbb{R}^{T_{\mathrm{txt}}\times d}$, formed by concatenation
$\mathbf{H}^{\mathrm{txt}}_{\ell}=[\mathbf{H}^{\mathrm{q}}_{\ell};\mathbf{H}^{\mathrm{aux}}_{\ell}]$ with
$T_{\mathrm{txt}}=T_q+T_{\mathrm{aux}}$.

We summarize the question by mean pooling,
\begin{equation}
\mathbf{q}_{\ell}=\mathrm{MeanPool}\!\left(\mathbf{H}^{\mathrm{q}}_{\ell}\right),
\end{equation}
and score each image token by cosine similarity to the question summary,
\begin{equation}
s_t=
\left\langle
\frac{\mathbf{q}_{\ell}}{\|\mathbf{q}_{\ell}\|_2},
\frac{\mathbf{H}^{\mathrm{img}}_{\ell}[t]}{\|\mathbf{H}^{\mathrm{img}}_{\ell}[t]\|_2}
\right\rangle.
\end{equation}

To isolate relevant visual context, we select the top-$K$ image tokens ($S$) with the highest similarity to the question. This focus ensures that the grounding correction is driven only by the most pertinent visual evidence.

We represent auxiliary text using the final decoder token ($\mathbf{H}^{\mathrm{aux}}_{\ell,\text{last}}$) rather than mean pooling. This choice is empirically superior, as the last token captures the full contextual summary and is positioned closest to the decision layer where option-redirecting bias occurs.

Using this representation, we compute the auxiliary--visual correction direction as
\begin{equation}
\Delta \mathbf{v}_{\ell} =
\frac{1}{K}
\sum_{i \in \mathcal{S}}
\left(
\mathbf{H}^{\mathrm{img}}_{\ell}[i]
-
\mathbf{H}^{\mathrm{aux}}_{\ell,\text{last}}
\right)
\end{equation}

The correction is uniformly applied to all text tokens, where
$j \in \{1,\dots,T_{\mathrm{txt}}\}$ indexes the text tokens:
\begin{equation}
\mathbf{H}^{\mathrm{txt}\prime}_{\ell}[j]
=
\mathbf{H}^{\mathrm{txt}}_{\ell}[j]
+
\beta_{\ell}\,\Delta \mathbf{v}_{\ell}
\end{equation}

The parameter $\beta_\ell$ is set to a small positive value to ensure that the update remains a residual adjustment and does not override the original representation.

DG operates as a self-adaptive mechanism that modulates the correction intensity based on the alignment between auxiliary text and visual evidence. By defining the correction vector $\Delta v_\ell$ as the discrepancy between these modalities, DG selectively intervenes: it triggers a robust correction when the auxiliary text is visually inconsistent (e.g., Option-Redirecting bias), but remains negligible ($\Delta v_\ell \approx 0$) when the text is faithful. This targeted design ensures that DG suppresses misleading signals while preserving the beneficial contributions of reliable auxiliary information.






\begin{figure*}
    \centering
    \includegraphics[width=1\linewidth]{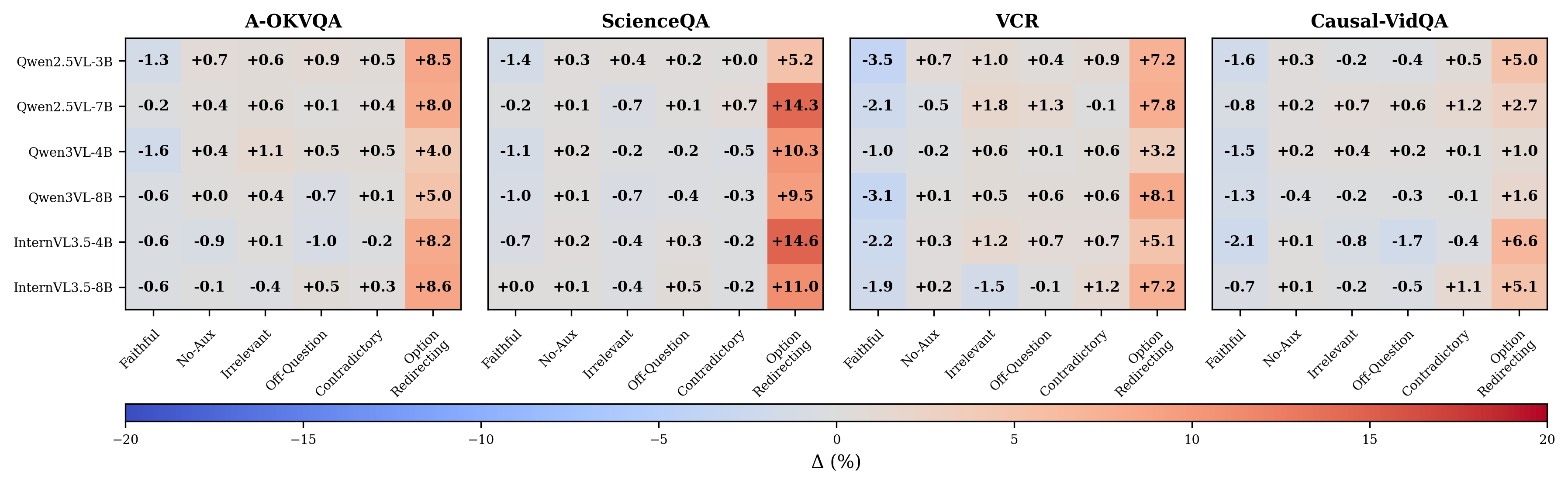}
    \caption{Accuracy change after NSS+DG relative to the frozen baseline. Improvements concentrate under Option-Redirecting text, with small changes in other conditions.}
    \Description{Heatmap of percentage-point accuracy changes from the combined NSS and DG intervention across models, datasets, and auxiliary-text conditions. The strongest positive changes are concentrated in the Option-Redirecting condition.}
    \label{fig:method}
\end{figure*}

\section{Experiments}

\subsection{Experimental Setup}

We evaluate frozen VLMs on multiple-choice VQA benchmarks under controlled variations of auxiliary text (Section~\ref{sec:ladder}). Experiments are conducted on A-OKVQA~\cite{schwenk2022okvqa}, ScienceQA~\cite{lu2022learn}, VCR~\cite{zellers2019recognition}, and Causal-VidQA (video-based)~\cite{li2022representation}, which span diverse multimodal reasoning scenarios and include human-annotated rationales.


For each dataset, we sample 1,000 valid evaluation instances and construct five ladder variants plus No-Aux. We use a fixed prompt, shared per-instance option permutation, and greedy decoding across all conditions. Complete sampling and inference details are provided in Appendix~\ref{app:prompt_inference}.


We evaluate representative open-source VLMs without fine-tuning to isolate decision-level biases in pretrained models: Qwen2.5VL (3B, 7B)~\cite{bai2025qwen2}, Qwen3VL (4B, 8B)~\cite{bai2025qwen3vltechnicalreport}, and InternVL3.5 (4B, 8B)~\cite{wang2025internvl3}. These models span multiple families and parameter scales, enabling us to examine whether auxiliary-text bias patterns are consistent across architectures and model capacities.

We use a fixed prompting template across datasets and auxiliary-text conditions to ensure consistent evaluation (see Appendix~\ref{app:prompt_inference} for details). Each input contains the image, question, auxiliary text, and shuffled multiple-choice options. We instruct the model to output \emph{only the option letter} (e.g., \texttt{A/B/C/D}) and perform greedy decoding (temperature $=0$) with a maximum generation length sufficient for a single option token. To prevent position-based shortcuts, we randomize the option order per instance using a fixed seed and reuse the same permutation across all auxiliary-text variants of that instance. Unless otherwise stated, we use a single fixed set of intervention hyperparameters per model--dataset pair, shared across all auxiliary-text conditions.

\subsection{Textual Reliability Ladder Results}

Figure~\ref{fig:textual_ladder} reports accuracy across datasets and model sizes under all auxiliary-text conditions, enabling direct comparison of model behavior across different levels of textual reliability.

\textbf{Faithful} auxiliary text consistently improves accuracy over \textbf{No-Aux} across benchmarks. This result indicates that frozen VLMs can effectively incorporate auxiliary text when it is jointly aligned with both the image and the question, validating the use of auxiliary text as a beneficial signal under reliable conditions.

Among the unreliable conditions, \textbf{Option-Redirecting} text produces the largest and most consistent accuracy drops across all models and datasets. By construction, this text remains question-relevant but conflicts with the image while implicitly supporting a specific distractor option, creating an option-targeted bias that can override visual grounding. To ensure that this effect is semantic rather than superficial, we control for potential artifacts during ladder construction: texts are style-matched, human-verified, and Option-Redirecting is token-length matched to Faithful. This regime uniquely combines question relevance, image inconsistency, and option-targeted support, directly corresponding to the failure mode isolated by the Textual Reliability Ladder.

In contrast, \textbf{Irrelevant}, \textbf{Off-Question}, and \textbf{Contradictory} text typically lead to smaller changes relative to \textbf{No-Aux}. This suggests that models rely more heavily on image--question cues when auxiliary text is weakly related, distracting, or inconsistent. Overall, these results demonstrate that auxiliary-text unreliability is not uniform in its impact. The most severe failures arise when auxiliary text introduces strong, option-targeted bias (\textbf{Option-Redirecting}), rather than when it is merely distracting or inconsistent (see Appendix~\ref{app:exp_ladder} for the full ladder results).

Scaling improves base performance---larger models achieve higher accuracy under \textbf{Faithful} and \textbf{No-Aux} conditions---but it does not monotonically improve robustness to misleading text. Under \textbf{Option-Redirecting}, performance drops remain substantial and can even increase with scale in some settings (e.g., Qwen2.5VL on ScienceQA and VCR). This suggests that larger models may more readily incorporate plausible, option-supporting textual evidence, even when it contradicts the visual input. Taken together, these findings show that improved model capacity does not necessarily translate to robustness under misleading auxiliary text, motivating inference-time interventions that mitigate option-targeted, text-driven decision bias.

\subsection{Two-stage Inference-Time Intervention}

We evaluate the proposed two-stage inference-time intervention under all auxiliary-text conditions. Figure~\ref{fig:method} reports accuracy differences relative to the corresponding frozen base models, enabling direct assessment of how the intervention affects model behavior across different auxiliary-text regimes.

The most pronounced improvements occur under \textbf{Option - Redirecting} auxiliary text: the two-stage intervention improves accuracy and reduces the frequency of selecting the redirected distractor $o_{\mathrm{redir}}$ (see Appendix~\ref{app:exp-option-redirecting} for a detailed analysis of redirected-distractor selection). In contrast, performance under \textbf{Faithful}, \textbf{Off-Question}, \textbf{Contradictory}, and \textbf{Irrelevant} conditions remains largely unchanged in Figure~\ref{fig:method}. This behavior is consistent with the design of the method, which primarily activates under strongly misleading auxiliary text while preserving useful or weak textual signals.

We observe a minor degradation under \textbf{Faithful} auxiliary text: a few model--dataset pairs show mild negative deltas. This behavior reflects a trade-off inherent to robustness-oriented steering. Mechanisms that suppress misleading auxiliary text can also attenuate beneficial text cues when the auxiliary text is already highly reliable. These regressions remain limited in magnitude and do not affect the overall conclusions, as the method remains largely conservative under \textbf{Faithful} text while delivering its largest gains in the harmful \textbf{Option-Redirecting} regime.

Stage~1 NSS improves robustness to superficial textual perturbations and yields modest, consistent gains across conditions (see Appendix~\ref{app:method_stage1} for the Stage~1-only ablation). Stage~2 DG is question-aware and is designed to counteract option-targeted misdirection, producing the largest improvements under \textbf{Option-Redirecting} (see Appendix~\ref{app:method_stage2} for the Stage~2-only ablation). The two stages play complementary roles. NSS provides a global bias correction that reduces overall sensitivity to textual perturbations, whereas DG operates at the instance level to resolve grounding conflicts when auxiliary text is question-relevant but visually inconsistent. This complementarity explains why improvements are concentrated under \textbf{Option-Redirecting} text (see Appendix~\ref{app:why_combine} for the combined-vs-single-stage comparison), while behavior under non-harmful conditions remains largely unchanged. The intervention yields limited gains under \textbf{Irrelevant} and \textbf{Off-Question} conditions, consistent with the absence of strong text-induced bias in these regimes.

\subsection{Ablation Study}

\begin{table}[t]
\centering
\small
\caption{Macro-averaged accuracy changes ($\Delta$, in percentage points) relative to frozen base models, aggregated across all model--dataset pairs. NSS introduces minimal shifts across conditions, while DG yields substantial improvements under \textbf{Option-Redirecting} text at the cost of slight degradation under \textbf{Faithful} text. The combined method (\textbf{NSS+DG}) achieves the best overall trade-off. Full regime-wise results are provided in Appendix~\ref{app:two_stage_appendix}. Non-OR includes No-Aux, Irrelevant, Off-Question, and Contradictory.}

\label{tab:stagewise_ablation_main}
\begin{tabular}{lccc}
\toprule
\rowcolor{gray!10}
Metric & NSS only & DG only & NSS + DG \\
\midrule
Avg. $\Delta$ Acc. (Faithful) & +0.11 & -1.50 & -1.30 \\
Avg. $\Delta$ Acc. (Option-Redirecting) & +0.24 & +6.72 & \textbf{+6.99} \\
Avg. $\Delta$ Acc. (Non-OR) & +0.06 & +0.07 & \textbf{+0.16} \\
\bottomrule
\end{tabular}
\end{table}

We ablate the contribution of each stage and the auxiliary-text representation used in DG. The results support the proposed coarse-to-fine design: NSS provides a conservative global stabilization, DG provides the primary correction under \textbf{Option-Redirecting} text, and the full two-stage method achieves the best overall trade-off between robustness and performance.

\paragraph{\textbf{Stage-wise ablation.}}

Table~\ref{tab:stagewise_ablation_main} confirms the intended division of labor: NSS is conservative but provides limited recovery under Option-Redirecting text, whereas DG delivers most of the targeted gain with a modest Faithful-text cost. Their combination achieves the strongest recovery while remaining stable across non-OR conditions.

\paragraph{\textbf{DG text representation.}}
We compare mean pooling with the last auxiliary-text token as the DG representation. The latter performs consistently better, especially under \textbf{Option-Redirecting} text, and is therefore used in all experiments

\section{Conclusion}

We study decision-level text bias in tri-modal multiple-choice VQA and show that the impact of unreliable auxiliary text is not uniform. The most severe failures arise when auxiliary text is question-aligned yet image-inconsistent and implicitly favors a specific incorrect option, leading to option-targeted errors. To analyze this behavior, we introduce the Textual Reliability Ladder, which identifies \textbf{Option-Redirecting} text as a dominant failure mode. We further propose a lightweight inference-time intervention that mitigates this bias while preserving performance under reliable auxiliary text. These results highlight the importance of fine-grained, decision-level analysis for improving robustness in tri-modal VQA.

\begin{acks}
This research was supported by The VNUHCM-University of Information Technology's Scientific Research Support Fund.
\end{acks}




\clearpage
\appendix
This supplementary material provides implementation and evaluation details that complement the main paper:
\begin{itemize}
    \item \textbf{Appendix A} describes the construction of the \textit{Textual Reliability Ladder}: prompt templates for generating each auxiliary-text condition, instance-wise style matching and token-length matching for Option-Redirecting vs Faithful. Token-length stats are reported and mandatory two-annotator human screening and verification. Additionally, we also report dataset-level auxiliary-text statistics and a worked example of all ladder variants.
    \item \textbf{Appendix B} explains why we evaluate \emph{frozen} VLMs and why fine-tuning is not a fair test for attributing decision-level text bias under ladder-controlled interventions.
    \item \textbf{Appendix C} provides full implementation details of the proposed two-stage inference-time intervention (NSS and DG), including steering-direction estimation, hyperparameter settings across model--dataset pairs, and design choices such as layer/token selection and auxiliary-text representation.
    \item \textbf{Appendix D} reports extended experimental results and analyses, including complete ladder tables for frozen baselines, stage-wise ablations (Only NSS / Only DG / NSS+DG), and redirect analysis that quantifies option-targeted collapse under \textbf{Option-Redirecting} text.
\end{itemize}

\section{Construction of the Textual Reliability Ladder}
\label{app:ladder_prompts}

This appendix describes how we construct the auxiliary-text variants for the \textbf{Textual Reliability Ladder}, including the prompt templates, augmentation pipeline, and filtering/verification rules. Across all conditions, we hold the image, question, and answer options fixed, and vary only the auxiliary text to enable controlled counterfactual comparisons.

We follow a two-step pipeline of \emph{constrained generation} and \emph{verification}. Unreliable variants are generated with \textbf{Gemini-2.5-Flash}~\cite{comanici2025gemini} using fixed prompt templates; we repeatedly sample candidates until they satisfy the ladder definition described in this appendix (e.g., style matching, token-length matching for Option-Redirecting vs Faithful and leakage/option-cue filters), followed by mandatory human verification. Faithful texts are taken from dataset-provided annotations when available; if a Faithful annotation violates any constraint, we discard the entire instance to preserve strict comparability across ladder conditions.

\noindent
\textbf{Ladder Conditions.} The following auxiliary-text regimes are constructed:

\begin{itemize}
    \item Faithful (dataset-provided)
    \item Off-Question
    \item Contradictory
    \item Option-Redirecting
    \item Irrelevant
    \item No-Aux (no auxiliary text)
\end{itemize}

\subsection{Faithful}
\label{app:Faithful}

Faithful texts are taken directly from dataset-provided. These texts are grounded in the image and relevant to the question, and typically support the correct answer option.


\subsection{Off-Question}
\label{app:off_question}

These texts are visually grounded but irrelevant to the question, and thus should not systematically support any option. We generate them by conditioning \textit{on the image} using the Off-Question prompt template in Table~\ref{tab:prompt-templates-off-question}.

\begin{table}[t]
\centering
\begin{tcolorbox}[promptbox, width=0.95\linewidth, title=\textbf{Prompt Templates for generating Off-Question text}]
You are an assistant that generates short texts for the provided images. Your text MUST be consistent with what is visually present in the images, but it MUST NOT contain any information that answers or helps answer the user's question and text MUST IGNORE and DONT MENTION to any answer options provided. If a visual detail might be relevant to answering the question, you should avoid mentioning it. Focus instead on secondary or generic visual aspects (such as general atmosphere, style, background elements) that are unrelated to the question. Never refer to the question in your text. Always respond with exactly one concise text and nothing else.

Question: \{question\}.

Caption MUST IGNORE and DON'T MENTION to any answer options: \{list\_answers\_str\}.

The visual inputs are either a single image or consecutive keyframes sampled from the same video scene.
Write ONE concise text that correctly describes some visible aspects of the images, but does NOT include any information that answers or helps answer the question above. The text must:
\begin{itemize}
    \item Be based only on what is visible in the images.
    \item Avoid any detail that could be used to answer the question.
    \item Not mention or refer to the question.
\end{itemize}
Output only the text, with no extra explanations.

\end{tcolorbox}
\caption{Prompt templates used to generate Off-Question text.}
\label{tab:prompt-templates-off-question}
\end{table}

\subsection{Contradictory}
\label{app:Contradictory}

These texts are generated solely from the question text and are not grounded in the image at all. They are logically consistent with the question but fabricated with respect to visual content. We generate them using the Contradictory prompt template in Table~\ref{tab:prompt-templates-Contradictory}.

\begin{table}[t]
\centering
\begin{tcolorbox}[promptbox, width=0.95\linewidth, title=\textbf{Prompt Templates for generating Contradictory text}]

You are an assistant that generates short captions ONLY from the text of the user's question. You NEVER see any images, you MUST NOT try to describe or guess the actual image content and caption MUST IGNORE and DONT MENTION to any answer options provided. Your caption should be logically consistent with the question. Always answer with a single concise caption, without any explanation.

Question: \{question\}.

Caption MUST IGNORE and DON'T MENTION to any answer options: \{list\_answers\_str\}.

Write ONE concise text that is on-topic with the question but deliberately introduces a plausible yet incorrect scenario that would contradict the true visual content. The caption must be non-targeted: avoid implying a specific answer choice, avoid providing a strong cue toward any single option, and do not include names/phrases that could match the answer options. Use only general wording derived from the question theme (no external facts). Do NOT mention images, photos, or pictures. Output only the caption.
\end{tcolorbox}
\caption{Prompt templates used to generate Contradictory text.}
\label{tab:prompt-templates-Contradictory}
\end{table}

\subsection{Option-Redirecting}
\label{app:option_redirecting}

Option-Redirecting texts are question-relevant, image-inconsistent, and intentionally crafted to implicitly support a specific incorrect answer option. We generate them using the Option-Redirecting prompt template in Table~\ref{tab:prompt-templates-Option-Redirecting}.

\begin{table}[t]
\centering
\begin{tcolorbox}[promptbox, width=0.95\linewidth, title=\textbf{Prompt Templates for generating Option-Redirecting text}]

You are an assistant that generates caption hallucinations redirected to a specific answer for a question.

Question: \{question\}.

Follow structure of this answer:

Answer redirected: \{answer\_correct\}

Caption generated: \{text\_faithful\}

Generate a caption for answer redirected: \{answer\_redirected\}.

Only output the caption without any other explanations and caption must be concise and follow the format.

\end{tcolorbox}
\caption{Prompt templates used to generate Option-Redirecting text.}
\label{tab:prompt-templates-Option-Redirecting}
\end{table}

The redirected option is sampled uniformly from incorrect answer choices. 

\begin{table*}[!t]
\centering
\renewcommand{\arraystretch}{1.1}
\setlength{\tabcolsep}{10pt} 
\small
\begin{tabular}{ll ccc cc cc}
\toprule
\multirow{2}{*}{\textbf{Model}} & \multirow{2}{*}{\textbf{Dataset}} & \multicolumn{4}{c}{\textbf{Noise-Stability Steering}} & \multicolumn{3}{c}{\textbf{Dynamic Grounding}} \\
\cmidrule(lr){3-6} \cmidrule(lr){7-9}
& & $\alpha$ image & $\alpha$ text & Layer range & N\_demos & $\beta$ DG & top $k$ & Layer range \\
\midrule

\multirow{4}{*}{Qwen2.5VL-3B} 
& Causal-VidQA & 0.5 & 0.3 & All & 30 & 0.4 & 20 & 19--33 \\
& ScienceQA & 0.5 & 0.4 & All & 50 & 0.5 & 7  & 19--33 \\
& A-OKVQA    & 0.4 & 0.4 & All & 50 & 0.4 & 7  & 19--33 \\
& VCR        & 0.5 & 0.5 & All & 50 & 0.4 & 7  & 19--33 \\
\midrule

\multirow{4}{*}{Qwen2.5VL-7B} 
& Causal-VidQA & 0.4 & 0.5 & All & 30 & 0.4 & 20 & 13--28 \\
& ScienceQA & 0.4 & 0.4 & All & 50 & 0.6 & 7  & 13--28 \\
& A-OKVQA    & 0.3 & 0.3 & All & 50 & 0.6 & 7  & 13--28 \\
& VCR        & 0.5 & 0.3 & All & 50 & 0.5 & 7  & 13--28 \\
\midrule

\multirow{4}{*}{Qwen3VL-4B} 
& Causal-VidQA & 0.5 & 0.3 & All & 30 & 0.2 & 20 & 9--34 \\
& ScienceQA & 0.5 & 0.3 & All & 50 & 0.4 & 7  & 9--34 \\
& A-OKVQA    & 0.3 & 0.3 & All & 50 & 0.4 & 7  & 9--34 \\
& VCR        & 0.5 & 0.3 & All & 50 & 0.2 & 7  & 9--34 \\
\midrule

\multirow{4}{*}{Qwen3VL-8B} 
& Causal-VidQA & 0.3 & 0.4 & All & 30 & 0.2 & 20 & 9--34 \\
& ScienceQA & 0.3 & 0.5 & All & 50 & 0.6 & 7  & 9--34 \\
& A-OKVQA    & 0.4 & 0.4 & All & 50 & 0.6 & 7  & 9--34 \\
& VCR        & 0.4 & 0.5 & All & 50 & 0.4 & 7  & 9--34 \\
\midrule

\multirow{4}{*}{InternVL3.5-4B} 
& Causal-VidQA & 0.3 & 0.1 & All & 15 & 0.4 & 20 & 9--34 \\
& ScienceQA & 0.5 & 0.1 & All & 20 & 0.5 & 7  & 9--34 \\
& A-OKVQA    & 0.4 & 0.1 & All & 20 & 0.3 & 7  & 9--34 \\
& VCR        & 0.3 & 0.1 & All & 20 & 0.3 & 7  & 9--34 \\
\midrule

\multirow{4}{*}{InternVL3.5-8B} 
& Causal-VidQA & 0.4 & 0.1 & All & 15 & 0.3 & 20 & 9--34 \\
& ScienceQA & 0.4 & 0.1 & All & 20 & 0.5 & 7  & 9--34 \\
& A-OKVQA    & 0.4 & 0.1 & All & 20 & 0.5 & 7  & 9--34 \\
& VCR        & 0.4 & 0.1 & All & 20 & 0.4 & 7  & 9--34 \\
\bottomrule
\end{tabular}

\caption{Best-performing hyperparameter configurations for our method (NSS + DG), used to produce the results in Figure~\ref{fig:method} (percentage accuracy gain over the base model) across models and datasets. The reported settings are selected via brute-force search over predefined ranges, aiming to identify configurations that consistently provide strong performance gains while maintaining stability across different models and datasets.}

\label{tab:hyperparameters}
\end{table*}

\subsection{Irrelevant Auxiliary Text}
\label{app:irrelevant}

Irrelevant auxiliary texts are sampled from WikiText~\cite{merity2016pointer} passages and filtered to avoid accidental overlap with the question or answer options (lexical-similarity, option-cue, and leakage filters). We do not enforce strict instance-wise length matching for Irrelevant; instead, we apply only a coarse token-budget bound to avoid degenerate verbosity/shortness and report token-length statistics in Appendix~\ref{app:aux-text-statis}.

\subsection{Human Verification}
\label{app:human_verification}

All constructed auxiliary texts undergo mandatory two-annotator human verification using a checklist aligned with the ladder definition:
(i) image-consistent vs.\ contradictory as required,
(ii) question-aligned vs.\ off-question behavior,
(iii) no option-targeted support when not intended,
(iv) for \textbf{Option-Redirecting}, the presence of option-targeted support without verbatim option leakage, and
(v) no answer leakage or invalid cues.
Ambiguous or non-compliant texts are removed or regenerated.

To control for length-based confounds in our key comparison, we enforce \emph{instance-wise} length matching for \textbf{Option-Redirecting} in \emph{token space} under the target model tokenizer:
$|L_{\text{OR}} - L_{\text{Faithful}}|\le \Delta_L$ (equivalently, $(1-\rho)L_{\text{Faithful}}\le L_{\text{OR}}\le (1+\rho)L_{\text{Faithful}}$).
For other conditions, we do not enforce strict matching; instead, we apply only a coarse length bound to avoid degenerate verbosity/shortness and report token-length statistics in Appendix~\ref{app:aux-text-statis}.

\paragraph{Faithful sample retention and target size.}
Since \textbf{Faithful} auxiliary text is sourced from the dataset, we do not rewrite it to satisfy our filtering or verification constraints, as doing so would introduce asymmetric editing across ladder conditions.
Instead, if the Faithful text for an instance violates the checklist (e.g., contains invalid cues or triggers answer leakage), we \emph{drop the entire instance} and continue sampling additional instances from the same evaluation split until we obtain exactly 1{,}000 \emph{valid} instances per dataset (fixed random seed), ensuring that all reported results are computed on the same target set size.

\paragraph{Independent annotation and adjudication.}
Verification is performed by two annotators working \emph{independently} with the same checklist.
An instance is accepted only if both annotators mark it as compliant.
Disagreements (including ``ambiguous'' cases) are resolved via adjudication: we discard the text and regenerate or remove the corresponding auxiliary variants until both annotators agree it satisfies the intended ladder condition.
This procedure is applied uniformly across all ladder conditions to maintain comparability and minimize subjective leakage or option-targeted artifacts.




\begin{table*}[htbp]
\centering
\small
\renewcommand{\arraystretch}{1.05}
\setlength{\tabcolsep}{3pt}
\begin{tabular}{l ccc ccc ccc ccc}
\toprule
\multirow{2}{*}{\textbf{Model}} 
& \multicolumn{3}{c}{\textbf{Causal-VidQA}} 
& \multicolumn{3}{c}{\textbf{ScienceQA}} 
& \multicolumn{3}{c}{\textbf{A-OKVQA}} 
& \multicolumn{3}{c}{\textbf{VCR}} \\
\cmidrule(lr){2-4} \cmidrule(lr){5-7} \cmidrule(lr){8-10} \cmidrule(lr){11-13}
& \shortstack{Unsteered\\\textit{(Frozen)}} 
& \shortstack{Two-stage\\\textit{(Ours)}} 
& \textbf{$\Delta(\%)$} 
& \shortstack{Unsteered\\\textit{(Frozen)}} 
& \shortstack{Two-stage\\\textit{(Ours)}} 
& \textbf{$\Delta(\%)$} 
& \shortstack{Unsteered\\\textit{(Frozen)}} 
& \shortstack{Two-stage\\\textit{(Ours)}} 
& \textbf{$\Delta(\%)$} 
& \shortstack{Unsteered\\\textit{(Frozen)}} 
& \shortstack{Two-stage\\\textit{(Ours)}} 
& \textbf{$\Delta(\%)$} \\
\midrule
Qwen2.5VL-3B   & 15.9 & 10.7 & 5.2  & 45.3 & 40.7 & 4.6  & 23.5 & 14.2 & 9.3  & 31.6 & 23.2 & 8.4 \\
Qwen2.5VL-7B   & 15.8 & 12.6 & 3.2  & 59.0 & 43.8 & 15.2 & 26.1 & 17.0 & 9.1  & 38.5 & 29.2 & 9.3 \\
Qwen3VL-4B     & 14.3 & 12.5 & 1.8  & 41.0 & 30.1 & 10.9 & 16.8 & 11.7 & 5.1  & 25.8 & 22.2 & 3.6 \\
Qwen3VL-8B     & 16.4 & 13.8 & 2.6  & 39.3 & 29.5 & 9.8  & 18.2 & 12.1 & 6.1  & 30.8 & 20.8 & 10.0 \\
InternVL3.5-4B & 30.0 & 20.6 & 9.4  & 63.8 & 48.6 & 15.2 & 46.8 & 37.9 & 8.9  & 47.4 & 39.6 & 7.8 \\
InternVL3.5-8B & 30.5 & 25.0 & 5.5  & 63.0 & 52.1 & 10.9 & 37.2 & 27.8 & 9.4  & 47.7 & 39.6 & 8.1 \\
\bottomrule
\end{tabular}
\caption{Redirect analysis across datasets. \emph{Redirect} reports the percentage of samples (out of 1{,}000) whose predictions match the redirected option under option-targeted biased text (lower is better). $\Delta(\%)$ denotes the absolute reduction in \emph{Redirect} (in percentage points) relative to the no-intervention baseline, computed as $\Delta = \mathrm{Redirect}~{\text{Unsteer (Frozen)}} - \mathrm{Redirect}~{\text{Two-stages (Our)}}$ (higher is better). Lower \emph{Redirect} indicates stronger robustness to option-redirecting textual bias.}

\label{tab:redirect-analysis}
\end{table*}

\subsection{Example: Textual Reliability Ladder}
\label{app:example_ladder}

We illustrate the \textbf{Textual Reliability Ladder} with a concrete instance (Table~\ref{tab:example-tri-modal-ladder}). The image, question, and answer options are held fixed, while only the auxiliary text is varied to instantiate each ladder condition.

\begin{table}[t]
\centering
\begin{tcolorbox}[promptbox, width=1\linewidth, title=\textbf{Example of the Textual Reliability Ladder}]
\paragraph{Image:} Figure ~\ref{fig:image_example}
\paragraph{Question.} \emph{What are the boys doing in the sand near the shoreline?}\\
\textit{Options:} (A) building castles \ \ (B) tunneling \ \ (C) fighting \ \ (D) eating

\paragraph{Auxiliary-text variants:}
\begin{itemize}
    \item \textbf{Faithful:} The boys are building a sand structure near the waterline.
    \item \textbf{Off-Question:} A cloudy day at a rocky beach, with seabirds overhead and an umbrella on a distant outcrop.
    \item \textbf{Contradictory:} The boys are flying a kite on the beach while standing on the wet sand.
    \item \textbf{Option-Redirecting:} Two boys are wrestling and shoving each other in the sand.
    \item \textbf{Irrelevant:} The WikiText corpus contains articles across many topics.
\end{itemize}
\end{tcolorbox}
\caption{\textbf{Worked example of the Textual Reliability Ladder.} 
We keep the image, question, and answer options fixed, and vary only the auxiliary text to instantiate the five ladder conditions (Faithful, Off-Question, Contradictory, Option-Redirecting, Irrelevant).}
\label{tab:example-tri-modal-ladder}
\end{table}

\subsection{Auxiliary Text Statistics}
\label{app:aux-text-statis}

\begin{figure}
    \centering
    \includegraphics[width=0.8\linewidth]{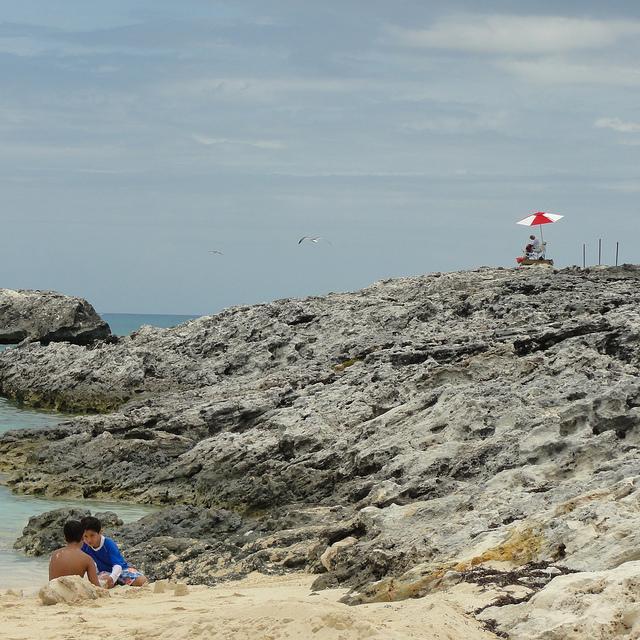}
    \caption{Image example}
    \Description{Rocky beach scene with two boys sitting near the water in the lower left and a person under a red-and-white umbrella on rocks in the distance.}
    \label{fig:image_example}
\end{figure}

Table~\ref{tab:aux_text_stats_transposed} reports dataset-level statistics of auxiliary texts used in our ladder evaluation.
For each dataset, we evaluate exactly 1{,}000 instances and summarize auxiliary-text \emph{token} lengths computed under the corresponding model tokenizer used at inference.
Cells report the per-instance token count (mean$\pm$std); values in parentheses denote the token-length difference relative to \textbf{Faithful},
$\Delta = L_{\text{cond}} - L_{\text{Faithful}}$ (in tokens).
We note that \textbf{Causal-VidQA} is a video QA benchmark where each instance is represented by multiple sampled frames as visual inputs.
Accordingly, the \textbf{Image} row reports the total number of frame inputs aggregated over the 1{,}000 evaluated instances (3{,}716),
whereas \textbf{A-OKVQA}, \textbf{ScienceQA}, and \textbf{VCR} are image-based datasets with a single image per instance.

Importantly, \textbf{Option-Redirecting} texts are closely length-matched to \textbf{Faithful} across datasets (small $|\Delta|$), which helps isolate option-targeted bias from superficial length effects.
In contrast, other conditions may exhibit larger length differences due to their definitions (e.g., \textbf{Contradictory} often forms short counter-statements, while \textbf{Irrelevant} may be longer and descriptive).

\begin{table*}[t]
\centering
\small
\setlength{\tabcolsep}{12pt}
\renewcommand{\arraystretch}{1.15}
\begin{tabular}{l c c c c}
\toprule
\textbf{Text Type} & \textbf{A-OKVQA} & \textbf{ScienceQA} & \textbf{VCR} & \textbf{Causal-VidQA} \\
\midrule
\textbf{Image} & 1000 & 1000 & 1000 & 3716 \\
\textbf{Faithful} & $12.72\!\pm\!7.07$ & $56.00\!\pm\!51.79$ & $20.20\!\pm\!10.22$ & $14.42\!\pm\!6.35$ \\
\textbf{Off-Question} & $20.34\!\pm\!4.55$ & $16.11\!\pm\!4.16$ & $12.92\!\pm\!2.84$ & $18.42\!\pm\!4.16$ \\
\textbf{Contradictory} & $5.66\!\pm\!2.01$ & $5.95\!\pm\!2.37$ & $8.86\!\pm\!3.30$ & $9.58\!\pm\!3.01$ \\
\textbf{Option-Redirecting} & $12.86\!\pm\!6.86$ & $53.33\!\pm\!49.00$ & $18.86\!\pm\!8.71$ & $14.22\!\pm\!6.74$ \\
\textbf{Irrelevant} & $25.26\!\pm\!12.88$ & $25.43\!\pm\!13.35$ & $24.80\!\pm\!12.89$ & $25.54\!\pm\!12.43$ \\
\bottomrule
\end{tabular}
\caption{Auxiliary-text length statistics for 1{,}000 evaluated instances per dataset. Cells report per-instance token lengths (mean$\pm$std). Strict instance-wise token-length matching is enforced only for \textbf{Option-Redirecting} relative to the corresponding \textbf{Faithful} text; other conditions are generated under fixed prompts with coarse length control.}
\label{tab:aux_text_stats_transposed}
\end{table*}


\noindent
In this example, the \textbf{Faithful} text accurately describes the relevant visual evidence (the boys building a sand structure), providing helpful context without introducing spurious details or favoring any incorrect option, consistent with the correct choice (option A: ``building castles'').
The \textbf{Option-Redirecting} text is \emph{question-aligned} but subtly biases the model toward a specific incorrect option (``fighting'') without using any option markers or verbatim option strings, whereas \textbf{Contradictory} introduces an incompatible activity that is not intentionally option-targeted.
\textbf{Off-Question} remains descriptive but uninformative for answering the question, and \textbf{Irrelevant} is unrelated background text. 
To isolate option-targeted bias from superficial length effects, Option-Redirecting is token-length matched to the corresponding Faithful text (within a fixed window), while other conditions are style-controlled but may vary in length (see Appendix~\ref{app:aux-text-statis}).

\paragraph{Quality control and removal criteria.}
To ensure that ladder effects arise from \emph{textual reliability} rather than superficial artifacts, we apply strict post-generation filtering to every auxiliary-text variant and \emph{remove or regenerate} any candidate that violates our constraints.
Concretely, we discard a text if it 
(i) contains any explicit multiple-choice markers or answer-revealing phrases (e.g., ``A/B/C/D'', ``(A)--(D)'', ``option~'C', ``the correct answer is''), 
(ii) copies or near-copies any answer-option string after normalization (including minor paraphrases that preserve distinctive option keywords), or 
(iii) exhibits \emph{answer leakage} by being disproportionately similar to a particular option compared with the others.
We further remove candidates that fail to satisfy the intended ladder semantics (e.g., an \textbf{Off-Question} description that mentions the boys' actions, or an \textbf{Option-Redirecting} text that becomes image-faithful), as well as low-quality generations such as repetitive phrasing, meta-instructions, or dataset-referential content.
Finally, we enforce per-instance token-length matching for \textbf{Option-Redirecting} vs.\ \textbf{Faithful}; for other conditions, we apply only a coarse token-budget bound to avoid degenerate verbosity/shortness.

\paragraph{Examples of discarded auxiliary text.}
Below are representative failure cases removed during filtering:
\begin{itemize}
    \item \textbf{Option-marker leakage:} ``The correct answer is option~(A).'' \ \ \emph{(explicit answer marker)}
    \item \textbf{Verbatim option copying:} ``They are building castles in the sand near the shoreline.'' \ \ \emph{(contains an option string)}
    \item \textbf{Near-copy / keyword leakage:} ``The boys are making sandcastles by the water.'' \ \ \emph{(near-duplicate of option~(A) via minor paraphrase)}
    \item \textbf{Semantic violation of condition:} \emph{Off-Question} candidate: ``Two boys dig a long tunnel in the sand.'' \ \ \emph{(becomes question-relevant)}
    \item \textbf{Meta / non-auxiliary text:} ``Select the best choice from the options based on the image.'' \ \ \emph{(instructional)}
    \item \textbf{Dataset-referential content:} ``This is a multiple-choice VQA sample with four options.'' \ \ \emph{(non-visual)}
    \item \textbf{Degenerate repetition:} ``Boys on the beach, boys on the beach, boys on the beach.'' \ \ \emph{(low-quality)}
\end{itemize}

\paragraph{Observation.}
Auxiliary-text length varies substantially across datasets (e.g., ScienceQA provides much longer Faithful rationales on average),
reflecting differences in annotation style and task format. Reporting these statistics contextualizes the robustness trends observed across ladder conditions.

\section{Why Fine-tuning Is Not a Fair Test of Textual Bias}
\label{app:why_no_finetune}

A core design choice of our study is to evaluate \emph{frozen} VLMs. This is not merely a computational preference, but a methodological requirement for making an objective claim about \emph{textual bias} induced by auxiliary text. Our goal is to isolate the \emph{causal effect} of auxiliary textual inputs (under the Textual Reliability Ladder in Section~\ref{sec:ladder}) on the model's final multiple-choice decision. If the model parameters are updated, performance changes can no longer be uniquely attributed to the reliability or type of the auxiliary text, because the model itself has changed.


Textual bias in our setting is defined as \emph{decision-level over-reliance} on auxiliary text, especially when the text is question-aligned but visually ungrounded (e.g., \textbf{Option-Redirecting}). Fine-tuning introduces multiple confounding factors that can mask or mimic bias reduction:
(i) \emph{general capability gains} (better instruction following, improved reasoning, better calibration),
(ii) \emph{dataset adaptation} (learning answer-option priors or annotation artifacts),
and (iii) \emph{prompt/format learning} (e.g., learning to output cleaner option tokens).
This concern is supported by prior work ~\cite{deng2503words}, which explicitly evaluates fine-tuning and shows that it tends to improve performance across \emph{all} textual conditions, including the \textbf{No-Aux} case. Such uniform gains do not necessarily imply reduced reliance on misleading text; rather, they are also consistent with a globally stronger model. Consequently, a fine-tuned model may appear ``more robust'' simply because it becomes better at the task overall, not because it has become less biased toward unreliable auxiliary descriptions.

In a ladder-style diagnosis, robustness is not equivalent to ignoring text. A model can reduce errors under \textbf{Option-Redirecting} by learning a blunt heuristic: \emph{always distrust auxiliary text}. While this may raise worst-case robustness, it can simultaneously harm performance under \textbf{Faithful} text where auxiliary information is genuinely useful. This failure mode is difficult to disentangle under fine-tuning because the training objective does not explicitly constrain the model to \emph{retain} beneficial text usage while \emph{selectively} resisting misleading cues. By contrast, frozen-model evaluation makes this trade-off transparent: if the model collapses under Option-Redirecting but succeeds under Faithful, the gap is attributable to the input reliability rather than to a shifted learned policy.

Our ladder evaluation is a controlled intervention: for each instance, we hold the image, question, and options fixed, and while controlling style and lexical shortcuts, and token-length matching Option-Redirecting to Faithful. This design is meaningful only when the model is fixed. Once fine-tuning is introduced, one must decide \emph{what} to fine-tune on (which datasets, which ladder conditions, what mixture ratios), and these choices can directly determine the apparent robustness. For example, including many Option-Redirecting samples in training may teach the model to discount auxiliary text in a dataset-specific manner, while excluding them yields a different behavior. Either way, the resulting comparison conflates the evaluation variable (text reliability) with the training variable (exposure and optimization), undermining the goal of an unbiased diagnosis.

Our method is designed as an \emph{inference-time intervention} that can be applied to off-the-shelf VLMs without retraining. This setting is practically important (e.g., when only black-box access or limited compute is available), and scientifically clean: keeping parameters frozen ensures that any behavioral change across ladder conditions is attributable to (a) the auxiliary text type, and (b) the explicit intervention we apply at inference time. Therefore, using frozen models provides a stricter and more interpretable testbed for studying textual bias and for validating mitigation mechanisms that do not rely on additional supervision or parameter updates.

In short, fine-tuning is not a fair test for \emph{measuring} textual bias under controlled ladder interventions, because it entangles robustness with general performance gains and training-dependent heuristics. Frozen-model evaluation is essential to attribute performance differences to auxiliary text reliability and to demonstrate that our inference-time approach specifically targets misleading textual influence rather than simply improving the model via parameter updates.

\section{Inference-Time Intervention}

\subsection{NSS Steering Vector Construction}
\label{app:demo_samples}

We illustrate the construction of NSS steering vectors with a concrete example from the dataset in Table~\ref{tab:nss-example1} and Image~\ref{fig:image_example}.

\begin{table}[t]
\centering
\small
\begin{tcolorbox}[width=0.47\textwidth]
\textbf{Example for NSS vector construction.}
\vspace{2pt}
\footnotesize
\begin{Verbatim}[breaklines=true, breakanywhere=true, fontsize=\footnotesize]
"correct_value": "A: building castles",
"incorrect_value": "C: fighting",
"question": "What are the boys doing in the sand near the 
shoreline?
Answer options:
A. building castles
B. tunneling
C. fighting
D. eating
Here is some additional information which are text
descriptions based on the image to assist you in 
answering the later question. Note, the information
could be irrelevant, missing some information or 
inaccurate, please use it with caution: 
Two boys are wrestling and shoving each other in the sand."
\end{Verbatim}
\end{tcolorbox}
\caption{A concrete example used to construct NSS steering vectors.}
\label{tab:nss-example1}
\end{table}

To construct NSS steering directions, we pair this instance with two answer continuations under an identical context:
(i) a correct continuation selecting option~A (building castles), and
(ii) a redirected incorrect continuation selecting option~C (fighting).

By computing the post-MLP hidden-state residuals between these two continuations—while holding the image, question, options, and auxiliary text fixed—we isolate representation differences associated with correct vs. incorrect answer trajectories. Aggregating such residuals across a small demonstration set and extracting the dominant principal component yields a global steering direction that captures systematic biases induced by misleading text. This direction is then applied uniformly at inference time to stabilize model behavior against textual noise.

\begin{figure*}
    \centering
    \includegraphics[width=1\linewidth]{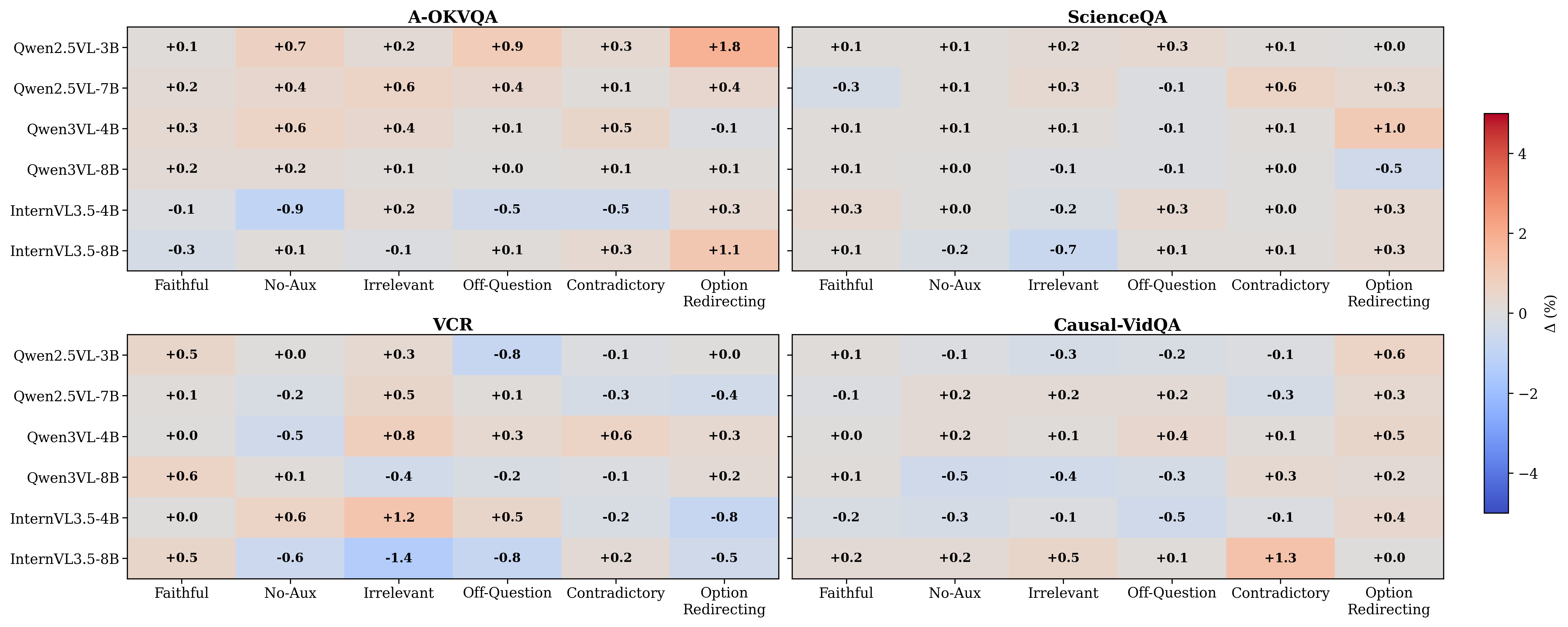}
    \caption{Heatmap of percentage accuracy improvement over the base models using only Stage 1 (NSS) under different auxiliary-text conditions.}
    \Description{Heatmap of percentage-point accuracy changes from Stage 1, Noise-Stability Steering, across models, datasets, and auxiliary-text conditions.}
    \label{fig:method_only_NSS}
\end{figure*}

\subsection{Stage 1: Noise-Stability Steering (NSS)}
\label{app:demo_hparams1}

To ensure a fair ladder comparison, we keep hyperparameters fixed across all textual conditions within each model–dataset pair. Table~\ref{tab:hyperparameters} summarizes the hyperparameter settings used for the proposed two-stage inference-time intervention across all evaluated models and datasets.
For Stage~1 NSS, we report the visual and textual steering coefficients ($\alpha_{\text{image}}$, $\alpha_{\text{text}}$), the decoder layer range where steering is applied, and the number of demonstrations used to estimate the steering directions.

The above construction yields a pair of layer-wise steering directions that are \emph{model-internal} and \emph{data-agnostic}: the visual direction captures the dominant drift under controlled visual corruption, while the textual direction captures the contrast between correct and misleading continuations under the same context. Importantly, both directions are estimated \emph{once} from a small demonstration set and remain fixed throughout evaluation, so NSS does not introduce any instance-specific tuning.

The goal of the textual residual is to capture an internal direction that distinguishes \emph{being driven to the correct option} from \emph{being attracted to a plausible but wrong option} under the \emph{same} visual--textual context. To this end, we construct two continuations for each demonstration: a correct answer $A^{+}$ and an incorrect answer $A^{-}$, while keeping the image, question, options, and auxiliary text fixed. The resulting representation difference therefore removes context effects and highlights a \emph{decision-separating} signal in the decoder space---i.e., the dominant direction along which the model’s hidden states shift when the selected option flips from wrong to right. In contrast, using only $A^{+}$ would mostly encode generic completion patterns or token identity, which are not aligned with our objective of reducing text-induced misselection. By explicitly including $A^{-}$ as a ``wrong-attractor'' trajectory, NSS can steer the decoder away from misleading option trajectories while preserving the ability to benefit from Faithful auxiliary text.

Across all models, NSS is applied uniformly across all layers. This design choice reflects the role of NSS as a global, static correction that reduces the model’s overall sensitivity to noisy or misleading textual signals, rather than resolving instance-specific grounding conflicts. Unlike Stage~2, NSS does not depend on the content of the current test sample and therefore provides a stable bias correction that is shared across inputs.

\subsection{Stage 2: Dynamic Grounding (DG)}
\label{app:demo_hparams2}

Stage~2 DG hyperparameters, shown in Table~\ref{tab:hyperparameters}, are more model- and dataset-dependent. We report the grounding strength $\beta_{\text{DG}}$, the number of question-relevant visual tokens $k$ selected for correction, and the decoder layer range where DG is applied.

DG is consistently applied to higher decoder layers, where semantic alignment between auxiliary text, visual evidence, and answer options has the strongest influence on final decision-making. In contrast to NSS, DG is instance-specific and question-aware, and is effective when auxiliary text conflicts with visually grounded evidence.

We observe clear dataset-dependent trends.
For video-based datasets (Causal-VidQA), a larger $k$ is used, reflecting the higher redundancy and temporal variation across frames. Selecting more visual tokens allows DG to aggregate consistent evidence across time.
For image-based datasets, smaller values of $k$ are sufficient, enabling DG to focus correction on the most discriminative visual regions without diluting the signal.

\paragraph{Auxiliary-text representation in DG.}
Dynamic Grounding (DG) requires a vector representation of the
auxiliary text to guide the adjustment of text-token hidden states.
We evaluated two alternatives: (i) mean pooling over all auxiliary-text
tokens and (ii) using the last auxiliary-text token.

Empirically, using the last token consistently yields stronger
performance across models and datasets, particularly under
Option-Redirecting auxiliary text. The final token in the
autoregressive decoder aggregates information from the preceding
auxiliary-text tokens through self-attention and lies closest to the
subsequent answer-generation step. As a result, it provides a more
direct signal for correcting option-level biases in the decoder.

Mean pooling, in contrast, averages across all tokens in the auxiliary
text and may dilute decision-relevant information with background or
descriptive tokens that are less related to the answer prediction.
Therefore, all DG results in this work use the last auxiliary-text
token as the auxiliary-text representation.

\section{Experiment}

\subsection{Experimental Setup}
\label{app:prompt_inference}

We apply a single fixed prompt template across all datasets, models, and textual conditions to avoid prompt-induced confounds:

\begin{table}[t]
\centering
\begin{tcolorbox}[promptbox, width=0.95\linewidth, title=\textbf{Prompt Templates for Evaluating}]
Question: \{question\}

Answer options:
\{options\}

Here are some additional information which are text descriptions based on the image to assist you for answering the later question. Note, the information could be irrelevant, missing some information or inaccurate, please use it with caution:
\{textual\}

Choose the CORRECT ID answer for the question below, you must response with the ID answer, don't include text.

\end{tcolorbox}
\caption{Prompt templates used to evaluate.}
\label{tab:prompt-templates-evaluate}
\end{table}

The template in Table~\ref{tab:prompt-templates-evaluate} explicitly cautions the model that the auxiliary text may be unreliable, ensuring that any observed errors reflect the model’s internal reliance on textual cues rather than prompt-induced misinstruction.

\begin{table*}[t]
\centering
\small
\renewcommand{\arraystretch}{1.1}
\setlength{\tabcolsep}{6pt}
\begin{tabular}{llcccccc}
\toprule
\textbf{Model} & \textbf{Dataset} & \textbf{Faithful} & \textbf{Contradictory} & \textbf{Off-Question} & \textbf{Option-Redirecting} & \textbf{Irrelevant} & \textbf{No-Aux} \\ 
\midrule
\multirow{4}{*}{Qwen2.5VL-3B} 
    & Causal-VidQA& 77.3 & 71.6 & 72.3 & 65.2 & 72.3 & 71.8 \\
    & ScienceQA & 92.9 & 75.1 & 75.1 & 49.3 & 74.9 & 73.1 \\
    & A-OKVQA    & 90.9 & 83.2 & 82.9 & 67.4 & 83.1 & 82.7 \\
    & VCR        & 83.1 & 69.7 & 69.8 & 52.4 & 69.1 & 69.1 \\
\midrule
\multirow{4}{*}{Qwen2.5VL-7B} 
    & Causal-VidQA& 77.9 & 71.7 & 73.3 & 65.3 & 72.6 & 72.1 \\
    & ScienceQA & 96.1 & 80.3 & 80.5 & 37.3 & 81.5 & 81.3 \\
    & A-OKVQA    & 91.8 & 84.8 & 84.9 & 66.3 & 85.0 & 85.2 \\
    & VCR        & 84.5 & 69.5 & 68.3 & 46.1 & 67.8 & 68.8 \\
\midrule
\multirow{4}{*}{Qwen3VL-4B} 
    & Causal-VidQA& 75.7 & 70.0 & 69.6 & 64.2 & 69.0 & 70.0 \\
    & ScienceQA & 96.8 & 83.2 & 82.2 & 55.7 & 82.7 & 82.8 \\
    & A-OKVQA    & 90.8 & 84.7 & 84.8 & 74.6 & 84.6 & 84.7 \\
    & VCR        & 77.1 & 65.7 & 66.5 & 56.4 & 66.7 & 66.9 \\
\midrule
\multirow{4}{*}{Qwen3VL-8B} 
    & Causal-VidQA& 77.5 & 70.9 & 71.6 & 64.6 & 71.2 & 71.2 \\
    & ScienceQA & 99.2 & 88.6 & 88.3 & 59.0 & 88.1 & 88.7 \\
    & A-OKVQA    & 92.4 & 87.8 & 88.0 & 75.5 & 87.7 & 88.3 \\
    & VCR        & 82.0 & 70.1 & 69.7 & 54.3 & 70.2 & 70.4 \\
\midrule
\multirow{4}{*}{InternVL3.5-4B} 
    & Causal-VidQA& 78.5 & 69.9 & 71.4 & 52.0 & 69.7 & 70.1 \\
    & ScienceQA & 98.3 & 87.9 & 86.9 & 35.5 & 87.9 & 88.6 \\
    & A-OKVQA    & 92.4 & 81.9 & 83.0 & 46.6 & 82.4 & 82.8 \\
    & VCR        & 82.4 & 65.8 & 64.8 & 38.7 & 64.6 & 66.3 \\
\midrule
\multirow{4}{*}{InternVL3.5-8B} 
    & Causal-VidQA& 80.6 & 71.1 & 71.7 & 52.9 & 71.2 & 72.2 \\
    & ScienceQA & 99.3 & 89.8 & 89.2 & 35.9 & 90.1 & 89.9 \\
    & A-OKVQA    & 93.4 & 85.2 & 84.7 & 56.3 & 85.8 & 85.2 \\
    & VCR        & 85.1 & 67.2 & 68.2 & 38.8 & 69.4 & 68.7 \\
\bottomrule
\end{tabular}

\caption{\textbf{Ladder evaluation on frozen VLMs (no intervention).} Accuracy (\%) under different auxiliary-text conditions (Faithful, Contradictory, Off-Question, Option-Redirecting, Irrelevant, and No-Aux), measured on frozen VLMs without applying our method or any parameter updates. Each dataset uses 1{,}000 samples.}

\label{tab:model_results}
\end{table*}

\begin{table*}[t]
\centering
\renewcommand{\arraystretch}{1.1}
\setlength{\tabcolsep}{6pt}
\small
\begin{tabular}{llcccccc}
\toprule
\textbf{Model} & \textbf{Dataset} & \textbf{Faithful} & \textbf{Contradictory} & \textbf{Off-Question} & \textbf{Option-Redirecting} & \textbf{Irrelevant} & \textbf{No-Aux} \\
\midrule
\multirow{4}{*}{Qwen2.5VL-3B}
    & Causal-VidQA & 77.4 & 71.5 & 72.1 & 65.8 & 72.0 & 71.7 \\
    & ScienceQA    & 93.0 & 75.2 & 75.4 & 49.3 & 75.1 & 73.2 \\
    & A-OKVQA      & 91.0 & 83.5 & 83.8 & 69.2 & 83.3 & 83.4 \\
    & VCR          & 83.6 & 69.6 & 69.0 & 52.4 & 69.4 & 69.1 \\
\midrule
\multirow{4}{*}{Qwen2.5VL-7B}
    & Causal-VidQA & 77.8 & 71.4 & 73.5 & 65.6 & 72.8 & 72.3 \\
    & ScienceQA    & 95.8 & 80.9 & 80.4 & 37.6 & 81.8 & 81.4 \\
    & A-OKVQA      & 92.0 & 84.9 & 85.3 & 66.7 & 85.6 & 85.6 \\
    & VCR          & 84.6 & 69.2 & 68.4 & 45.7 & 68.3 & 68.6 \\
\midrule
\multirow{4}{*}{Qwen3VL-4B}
    & Causal-VidQA & 75.7 & 70.1 & 70.0 & 64.7 & 69.1 & 70.2 \\
    & ScienceQA    & 96.9 & 83.3 & 82.1 & 56.7 & 82.8 & 82.9 \\
    & A-OKVQA      & 91.1 & 85.2 & 84.9 & 74.5 & 85.0 & 85.3 \\
    & VCR          & 77.1 & 66.3 & 66.8 & 56.7 & 67.5 & 66.4 \\
\midrule
\multirow{4}{*}{Qwen3VL-8B}
    & Causal-VidQA & 77.6 & 71.2 & 71.3 & 64.8 & 70.8 & 70.7 \\
    & ScienceQA    & 99.3 & 88.6 & 88.2 & 58.5 & 88.0 & 88.7 \\
    & A-OKVQA      & 92.6 & 87.9 & 88.0 & 75.6 & 87.8 & 88.5 \\
    & VCR          & 82.6 & 70.0 & 69.5 & 54.5 & 69.8 & 70.5 \\
\midrule
\multirow{4}{*}{InternVL3.5-4B}
    & Causal-VidQA & 78.3 & 69.8 & 70.9 & 52.4 & 69.6 & 69.8 \\
    & ScienceQA    & 98.6 & 87.9 & 87.2 & 35.8 & 87.7 & 88.6 \\
    & A-OKVQA      & 92.3 & 81.4 & 82.5 & 46.9 & 82.6 & 81.9 \\
    & VCR          & 82.4 & 65.6 & 65.3 & 37.9 & 65.8 & 66.9 \\
\midrule
\multirow{4}{*}{InternVL3.5-8B}
    & Causal-VidQA & 80.8 & 72.4 & 71.8 & 52.9 & 71.7 & 72.4 \\
    & ScienceQA    & 99.4 & 89.9 & 89.3 & 36.2 & 89.4 & 89.7 \\
    & A-OKVQA      & 93.1 & 85.5 & 84.8 & 57.4 & 85.7 & 85.3 \\
    & VCR          & 85.6 & 67.4 & 67.4 & 38.3 & 68.0 & 68.1 \\
\bottomrule
\end{tabular}
\caption{\textbf{Only NSS on frozen VLMs.} Ladder results reported as accuracy (\%) under different auxiliary-text conditions (Faithful, Contradictory, Off-Question, Option-Redirecting, Irrelevant, and No-Aux). Only NSS is applied at inference time (no training or parameter updates). Each dataset uses 1{,}000 samples.}
\label{tab:evaluation_only_NSS}
\end{table*}

\begin{table*}[t]
\centering
\renewcommand{\arraystretch}{1.1}
\setlength{\tabcolsep}{6pt}
\small
\begin{tabular}{llcccccc}
\toprule
\textbf{Model} & \textbf{Dataset} & \textbf{Faithful} & \textbf{Contradictory} & \textbf{Off-Question} & \textbf{Option-Redirecting} & \textbf{Irrelevant} & \textbf{No-Aux} \\
\midrule
\multirow{4}{*}{Qwen2.5VL-3B}
    & Causal-VidQA & 75.3 & 71.4 & 72.0 & 70.1 & 72.1 & 71.8 \\
    & ScienceQA    & 91.3 & 74.8 & 75.3 & 54.6 & 74.4 & 72.8 \\
    & A-OKVQA      & 89.1 & 83.1 & 83.8 & 75.1 & 83.5 & 83.0 \\
    & VCR          & 79.6 & 70.5 & 69.9 & 59.3 & 70.9 & 69.6 \\
\midrule
\multirow{4}{*}{Qwen2.5VL-7B}
    & Causal-VidQA & 77.0 & 72.6 & 73.7 & 67.2 & 73.1 & 72.1 \\
    & ScienceQA    & 96.0 & 80.9 & 80.3 & 51.3 & 80.6 & 81.3 \\
    & A-OKVQA      & 91.3 & 85.1 & 84.8 & 73.9 & 85.3 & 85.5 \\
    & VCR          & 81.6 & 69.7 & 69.4 & 54.1 & 69.6 & 69.0 \\
\midrule
\multirow{4}{*}{Qwen3VL-4B}
    & Causal-VidQA & 74.1 & 69.6 & 69.8 & 65.1 & 69.4 & 70.1 \\
    & ScienceQA    & 95.7 & 82.4 & 82.1 & 64.5 & 82.4 & 82.8 \\
    & A-OKVQA      & 89.1 & 84.8 & 85.6 & 78.2 & 85.6 & 84.8 \\
    & VCR          & 75.6 & 66.3 & 66.9 & 59.5 & 67.1 & 67.4 \\
\midrule
\multirow{4}{*}{Qwen3VL-8B}
    & Causal-VidQA & 76.0 & 71.0 & 71.3 & 66.3 & 71.3 & 71.1 \\
    & ScienceQA    & 98.2 & 87.8 & 87.7 & 68.5 & 87.3 & 88.7 \\
    & A-OKVQA      & 91.6 & 87.6 & 87.6 & 80.6 & 87.7 & 88.3 \\
    & VCR          & 79.0 & 70.3 & 70.0 & 61.9 & 70.0 & 70.4 \\
\midrule
\multirow{4}{*}{InternVL3.5-4B}
    & Causal-VidQA & 77.0 & 70.4 & 70.3 & 58.8 & 69.7 & 70.1 \\
    & ScienceQA    & 97.6 & 87.6 & 87.1 & 50.6 & 87.2 & 88.7 \\
    & A-OKVQA      & 92.0 & 81.9 & 82.4 & 54.1 & 82.8 & 82.8 \\
    & VCR          & 79.5 & 65.9 & 64.9 & 43.7 & 65.5 & 66.8 \\
\midrule
\multirow{4}{*}{InternVL3.5-8B}
    & Causal-VidQA & 78.8 & 70.9 & 71.0 & 56.0 & 70.0 & 72.2 \\
    & ScienceQA    & 99.3 & 89.4 & 89.7 & 47.5 & 89.4 & 90.2 \\
    & A-OKVQA      & 92.7 & 85.1 & 85.5 & 64.7 & 85.4 & 85.2 \\
    & VCR          & 82.7 & 67.9 & 67.5 & 46.0 & 68.9 & 68.7 \\
\bottomrule
\end{tabular}
\caption{\textbf{Only DG on frozen VLMs.} Ladder results reported as accuracy (\%) under different auxiliary-text conditions (Faithful, Contradictory, Off-Question, Option-Redirecting, Irrelevant, and No-Aux). Only DG is applied at inference time (no training or parameter updates). Each dataset uses 1{,}000 samples.}
\label{tab:evaluation_only_DG}
\end{table*}

\begin{table*}[t]
\centering
\renewcommand{\arraystretch}{1.1}
\setlength{\tabcolsep}{6pt}
\small
\begin{tabular}{llcccccc}
\toprule
\textbf{Model} & \textbf{Dataset} & \textbf{Faithful} & \textbf{Contradictory} & \textbf{Off-Question} & \textbf{Option-Redirecting} & \textbf{Irrelevant} & \textbf{No-Aux} \\
\midrule
\multirow{4}{*}{Qwen2.5VL-3B}
    & Causal-VidQA & 75.7 & 72.1 & 71.9 & 70.2 & 72.1 & 72.1 \\
    & ScienceQA    & 91.5 & 75.1 & 75.3 & 54.5 & 75.3 & 73.4 \\
    & A-OKVQA      & 89.6 & 83.7 & 83.8 & 75.9 & 83.7 & 83.4 \\
    & VCR          & 79.6 & 70.6 & 70.2 & 59.6 & 70.1 & 69.8 \\
\midrule
\multirow{4}{*}{Qwen2.5VL-7B}
    & Causal-VidQA & 77.1 & 72.9 & 73.9 & 68.0 & 73.3 & 72.3 \\
    & ScienceQA    & 95.9 & 81.0 & 80.6 & 51.6 & 80.8 & 81.4 \\
    & A-OKVQA      & 91.6 & 85.2 & 85.0 & 74.3 & 85.6 & 85.6 \\
    & VCR          & 82.4 & 69.4 & 69.6 & 53.9 & 69.6 & 68.3 \\
\midrule
\multirow{4}{*}{Qwen3VL-4B}
    & Causal-VidQA & 74.2 & 70.1 & 69.8 & 65.2 & 69.4 & 70.2 \\
    & ScienceQA    & 95.7 & 82.7 & 82.0 & 66.0 & 82.5 & 83.0 \\
    & A-OKVQA      & 89.2 & 85.2 & 85.3 & 78.6 & 85.7 & 85.1 \\
    & VCR          & 76.1 & 66.3 & 66.6 & 59.6 & 67.3 & 66.7 \\
\midrule
\multirow{4}{*}{Qwen3VL-8B}
    & Causal-VidQA & 76.2 & 70.8 & 71.3 & 66.2 & 71.0 & 70.8 \\
    & ScienceQA    & 98.2 & 88.3 & 87.9 & 68.5 & 87.4 & 88.8 \\
    & A-OKVQA      & 91.8 & 87.9 & 87.3 & 80.5 & 88.1 & 88.3 \\
    & VCR          & 78.9 & 70.7 & 70.3 & 62.4 & 70.7 & 70.5 \\
\midrule
\multirow{4}{*}{InternVL3.5-4B}
    & Causal-VidQA & 76.4 & 69.5 & 69.7 & 58.6 & 68.9 & 70.2 \\
    & ScienceQA    & 97.6 & 87.7 & 87.2 & 50.1 & 87.5 & 88.8 \\
    & A-OKVQA      & 91.8 & 81.7 & 82.0 & 54.8 & 82.5 & 81.9 \\
    & VCR          & 80.2 & 66.5 & 65.5 & 43.8 & 65.8 & 66.6 \\
\midrule
\multirow{4}{*}{InternVL3.5-8B}
    & Causal-VidQA & 79.9 & 72.2 & 71.2 & 58.0 & 71.0 & 72.3 \\
    & ScienceQA    & 99.3 & 89.6 & 89.7 & 46.9 & 89.7 & 90.0 \\
    & A-OKVQA      & 92.8 & 85.5 & 85.2 & 64.9 & 85.4 & 85.1 \\
    & VCR          & 83.2 & 68.4 & 68.1 & 46.0 & 67.9 & 68.9 \\
\bottomrule
\end{tabular}
\caption{\textbf{Combined intervention on frozen VLMs.} Ladder results reported as accuracy (\%) under different auxiliary-text conditions (Faithful, Contradictory, Off-Question, Option-Redirecting, Irrelevant, and No-Aux). We apply the combined inference-time components (DG + NSS) without any training or parameter updates. Each dataset uses 1{,}000 samples.}
\label{tab:combine-result}
\end{table*}

\subsection{Textual Reliability Ladder Results}
\label{app:exp_ladder}
Table~\ref{tab:model_results} reports ladder performance for \emph{frozen} VLMs without any inference-time intervention. We observe a clear reliability gradient: accuracy is highest with \textbf{Faithful} auxiliary text and degrades as text becomes less reliable, with \textbf{Option-Redirecting} producing the largest drops. These results indicate that when auxiliary text is fluent and question-aligned, current VLMs can exhibit strong \emph{decision-level} reliance on it.

We use No-Aux as the text-free baseline to quantify the \emph{net} effect of adding auxiliary text (help or harm), and Faithful as the reliability upper bound to characterize the ladder trend.

Across models and datasets, Faithful consistently outperforms No-Aux, demonstrating that image-consistent auxiliary text provides useful contextual evidence. In contrast, Irrelevant remains close to No-Aux, suggesting that purely unrelated text neither systematically helps nor harms.

When auxiliary text is Off-Question (visually grounded but not question-relevant), performance generally drops relative to Faithful and is often close to or slightly below No-Aux, consistent with mild distraction. Contradictory text further reduces accuracy, demonstrating that plausible but image-inconsistent statements can mislead VLMs even without option cues.

Option-Redirecting yields the most severe degradation across models and datasets, and typically underperforms No-Aux, evidencing substantial net harm from biased auxiliary text. The effect is especially pronounced on ScienceQA and VCR: for example, Qwen2.5VL-7B drops from $96.1\%$ (Faithful) to $37.3\%$ (Option-Redirecting) on ScienceQA and from $84.5\%$ to $46.1\%$ on VCR; InternVL3.5-4B shows similar drops ($98.3\%\rightarrow35.5\%$ on ScienceQA; $82.4\%\rightarrow38.7\%$ on VCR). On A-OKVQA, the degradation remains substantial but smaller (e.g., Qwen2.5VL-3B: $90.9\%\rightarrow67.4\%$), while Causal-VidQA exhibits the smallest drop (e.g., $77.3\%\rightarrow65.2\%$ for Qwen2.5VL-3B), consistent with stronger constraints from direct visual evidence.

\paragraph{Model size helps in clean settings but does not remove text bias.}
Larger backbones often improve \textbf{Faithful} accuracy (e.g., 7B/8B variants are generally stronger than 3B/4B),
yet they remain vulnerable to Option-Redirecting bias.
Moreover, susceptibility varies by model family: for ScienceQA, Qwen3VL-8B retains $59.0\%$ under Option-Redirecting,
whereas InternVL3.5-8B falls to $35.9\%$, indicating that architectural/training differences affect how strongly
auxiliary text dominates the final decision.

The ladder results establish a clear reliability gradient and identify Option-Redirecting as the
dominant failure mode for frozen VLMs, motivating our two-stage inference-time intervention that specifically targets
Option-Redirecting text bias while maintaining stability under other text conditions.

\subsubsection{Why We Do Not Compare Against Additional Baselines}
\label{app:no_compare_baselines}

Our goal is to \emph{diagnose} and \emph{mitigate} decision-level text bias induced by auxiliary text under a controlled Textual Reliability Ladder, while keeping the underlying VLM \emph{frozen}. This setting imposes specific requirements on what constitutes a fair and meaningful baseline. Below we clarify why we do not include several common categories of approaches.

We do not compare against fine-tuning baselines because updating model parameters confounds ladder attribution: performance changes may reflect general capability gains or training-induced heuristics rather than reduced reliance on unreliable auxiliary text (see Appendix~\ref{app:why_no_finetune}). Therefore, we restrict our study to frozen models to ensure a fair and interpretable ladder comparison.

Most prior ``debiasing'' work in VQA focuses on \emph{dataset-level language priors} (e.g., question-only shortcuts) and typically relies on architectural changes, auxiliary branches, counterfactual training, or additional supervision. These methods are not directly comparable to our setting, where the bias source is \emph{externally supplied auxiliary text} whose reliability is systematically manipulated (Faithful / Contradictory / Option-Redirecting / etc.). In particular, our failure mode is \emph{Option-Redirecting} misselection driven by question-aligned but visually ungrounded text, which is structurally different from classic question-only priors.

\paragraph{Test-time approaches for VLMs ``language-prior'' bias are not aligned with our ladder protocol.}
There exist training-free/post-hoc strategies that calibrate VLMs outputs using counterfactual inputs such as removing or corrupting the image, with the goal of reducing reliance on the underlying LLM prior. However, these methods are designed around an \emph{image-missing or image-incongruent} baseline and primarily correct for \emph{vision-ignoring} behavior, rather than for \emph{auxiliary-text reliability} and Option-Redirecting redirection under a fixed image. Applying such calibration would also change the evaluation protocol by introducing additional counterfactual forward passes and assumptions (e.g., ``uniform'' outputs when the image is absent), which are orthogonal to our ladder interventions and may penalize cases where Faithful auxiliary text is genuinely informative. For these reasons, we do not treat them as direct baselines for ladder robustness to misleading auxiliary text.

Prompting is known to affect modality preference and text bias. Prior work~\cite{deng2503words} systematically evaluates prompt factors (e.g., instruction phrasing, token order) and demonstrates that prompt adjustments alone do not provide a consistent, model-agnostic solution to blind faith in text. Our evaluation therefore adopts a fixed prompting template that follows prior best practice and is kept \emph{identical} across all ladder conditions, models, and datasets. Since prompt engineering is not our target variable and has been analyzed in depth in prior work, additional prompt baselines would not be informative for isolating the effect of auxiliary-text reliability or for validating our inference-time mechanism.

Overall, we focus on the most controlled and interpretable comparison: frozen VLMs under a fixed prompt, where only the auxiliary-text type (and our inference-time intervention) changes. This design enables an unbiased ladder diagnosis and a clean attribution of robustness gains to the proposed two-stage method rather than to training-dependent confounds.

\subsection{Two-stage Inference-Time Intervention}
\label{app:two_stage_appendix}

\subsubsection{Only Noise-Stability Steering (NSS) (Stage 1).}
\label{app:method_stage1}

We ablate our two-stage intervention by enabling only Stage~1, \emph{NSS}, while disabling Stage~2 (DG).
Recall that NSS is a \emph{global, static} activation-level shift: the visual stream is made more noise-stable by subtracting a visual direction,
and the text decoder is weakly biased toward correct continuations by adding a textual direction. Once the per-layer directions are computed
offline, NSS is applied uniformly via forward hooks and requires \emph{no extra forward passes} or per-instance adaptation at test time.

Across models, datasets, and ladder conditions, NSS yields a modest average improvement of about $+0.10$ accuracy points
(i.e., \emph{percentage points}) and rarely causes large regressions (Figure~\ref{fig:method_only_NSS}, Table~\ref{tab:evaluation_only_NSS}).
This is consistent with NSS being option-agnostic and designed primarily as a \emph{stability prior} rather than a fine-grained grounding fix.
Notably, performance under \textbf{Faithful} auxiliary text remains essentially unchanged (average $+0.11\%$), indicating that NSS does not
sacrifice accuracy when the auxiliary text is reliable. We report macro-averaged gains. For each condition, we average $\Delta$Accuracy (NSS minus Frozen) over all model--dataset pairs (24 cells).
For the overall effect across conditions, we additionally average over all ladder conditions (144 cells total).

The largest average gains occur under \textbf{Option-Redirecting} text (average $+0.24\%$), where the model is most vulnerable to
question-aligned but image-inconsistent evidence.
For instance, on A-OKVQA with Qwen2.5VL-3B, NSS improves Option-Redirecting accuracy from $67.4\%$ to $69.2\%$ (+1.8\%),
while also slightly improving Faithful accuracy ($90.9\%\rightarrow 91.0\%$).
At the dataset level, this pattern is most pronounced on A-OKVQA (Option-Redirecting average $+0.60\%$), suggesting that NSS can
partially reduce the model's tendency to over-trust semantically plausible textual cues in knowledge-heavy VQA.

\subsubsection{Only Dynamic Grounding (DG) (Stage 2).}
\label{app:method_stage2}

We next isolate Stage~2 by enabling only \emph{(DG)} on frozen VLMs, while disabling Stage~1 (NSS).
Unlike NSS (a global, static shift), DG is \emph{instance-dependent}: it uses the current image--question context to
adjust how auxiliary text is incorporated at inference time, aiming to suppress text-driven shortcuts when the text
is misleading.

\begin{figure*}[!t]
    \centering \includegraphics[width=1\linewidth]{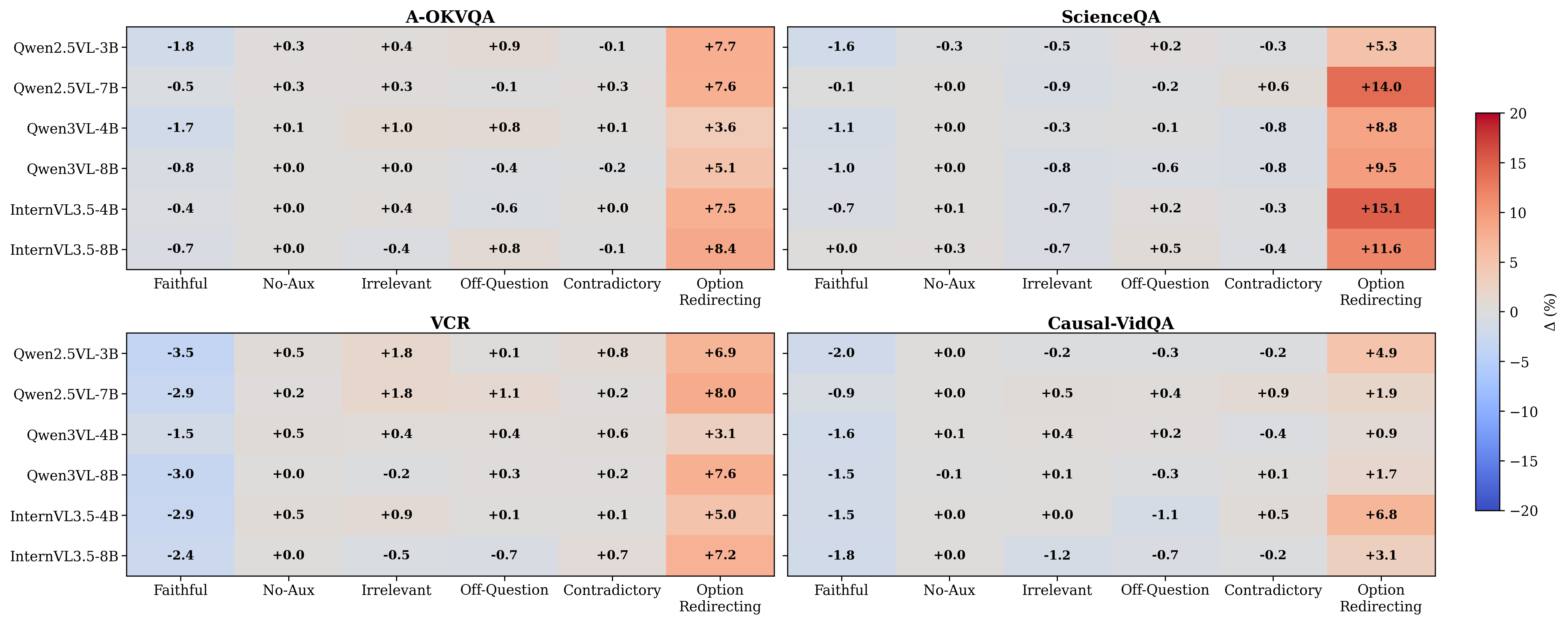}
    \caption{Heatmap of percentage accuracy improvement over the base models using only Stage 2 (DG) under different auxiliary-text conditions. Gains are largest under Option-Redirecting textual descriptions, while remaining stable across other auxiliary-text settings.}
    \Description{Heatmap of percentage-point accuracy changes from Stage 2, Dynamic Grounding, across models, datasets, and auxiliary-text conditions. The largest gains appear under Option-Redirecting text.}
    \label{fig:method_only_DG}
\end{figure*}

The ablation in Figure~\ref{fig:method_only_DG} shows that DG delivers substantially larger gains than NSS precisely in the failure mode we target: \textbf{Option-Redirecting} auxiliary text.
Across all four datasets, improvements are consistently positive and often large (up to double digits), indicating that DG effectively counters Option-Redirecting misdirection.
For example, on ScienceQA, DG yields strong gains across model families (e.g., +14.0\% for Qwen2.5VL-7B and +15.1\% for InternVL3.5-4B under Option-Redirecting; Figure~\ref{fig:method_only_DG}). Similar trends appear on A-OKVQA (e.g., +7.7\% on Qwen2.5VL-3B and +8.4\% on InternVL3.5-8B) and on VCR (e.g., +6.9\% on Qwen2.5VL-3B and +8.0\% on Qwen2.5VL-7B).
These results support the intended role of DG: when auxiliary text is question-aligned but option-biased, DG can
re-anchor the decision to question-relevant visual evidence, preventing the model from following the redirected
textual narrative.

Under \textbf{No-Aux}, \textbf{Irrelevant}, and \textbf{Off-Question} conditions, the changes are generally small
(near zero for most model--dataset pairs; Figure~\ref{fig:method_only_DG} and Table~\ref{tab:evaluation_only_DG}).
This suggests DG does not introduce large distributional shifts when the auxiliary text is uninformative, and that
its benefit primarily emerges when there is a strong text-induced bias to correct.

A consistent pattern in Figure~\ref{fig:method_only_DG} is that DG can be mildly to moderately negative under \textbf{Faithful} text
(e.g., around $-0.4\%$ to $-1.8\%$ on A-OKVQA/ScienceQA, and up to $-3.5\%$ on VCR for Qwen2.5VL-3B).
This is expected from an intervention whose goal is to \emph{discount} potentially unreliable text: when auxiliary
text is actually correct and helpful, aggressively reweighting or suppressing its influence can remove useful
signal and reduce accuracy. Therefore, DG alone is not a universally safe default.

Overall, the Only DG ablation indicates that DG is the \emph{primary mechanism} for mitigating Option-Redirecting text
bias, producing large and consistent improvements under Option-Redirecting auxiliary text while remaining mostly
stable when text is weakly informative. However, DG alone may degrade performance when the auxiliary text is
Faithful, motivating the two-stage design: NSS provides a conservative stabilization backbone, and DG supplies
instance-aware correction when the auxiliary text is likely to be misleading.

\subsubsection{Combine Two-stage}
\label{app:combine-two-stage}

Finally, we apply the full two-stage inference-time intervention by combining \emph{NSS} (Stage~1) and \emph{DG} (Stage~2) on frozen VLMs.
This setting reflects our intended use case: NSS provides a lightweight, model-wide stabilization of internal representations,
while DG performs \emph{instance-dependent} re-grounding to suppress auxiliary-text shortcuts when the text is misleading.
Both components are applied at test time only, requiring no training or parameter updates.

Figure~\ref{fig:method} shows a clear and consistent pattern: the combined method yields its strongest improvements under
Option-Redirecting auxiliary text across all datasets and model families.
The gains are often substantial, reaching up to $+14.6\%$ points on ScienceQA (e.g., InternVL3.5-4B) and $+14.3\%$ on ScienceQA (Qwen2.5VL-7B), while also being large on A-OKVQA (up to $+8.6\%$), VCR (up to $+8.1\%$), and Causal-VidQA (up to $+6.6\%$).
These results match the motivation of our approach: when the auxiliary text is question-aligned but intentionally pushes the model toward a wrong option, the two-stage intervention can effectively reduce option-targeted text bias and recover a significant portion of the lost accuracy (Table~\ref{tab:combine-result}).

\begin{figure*}
   \centering
   \includegraphics[width=1\linewidth]{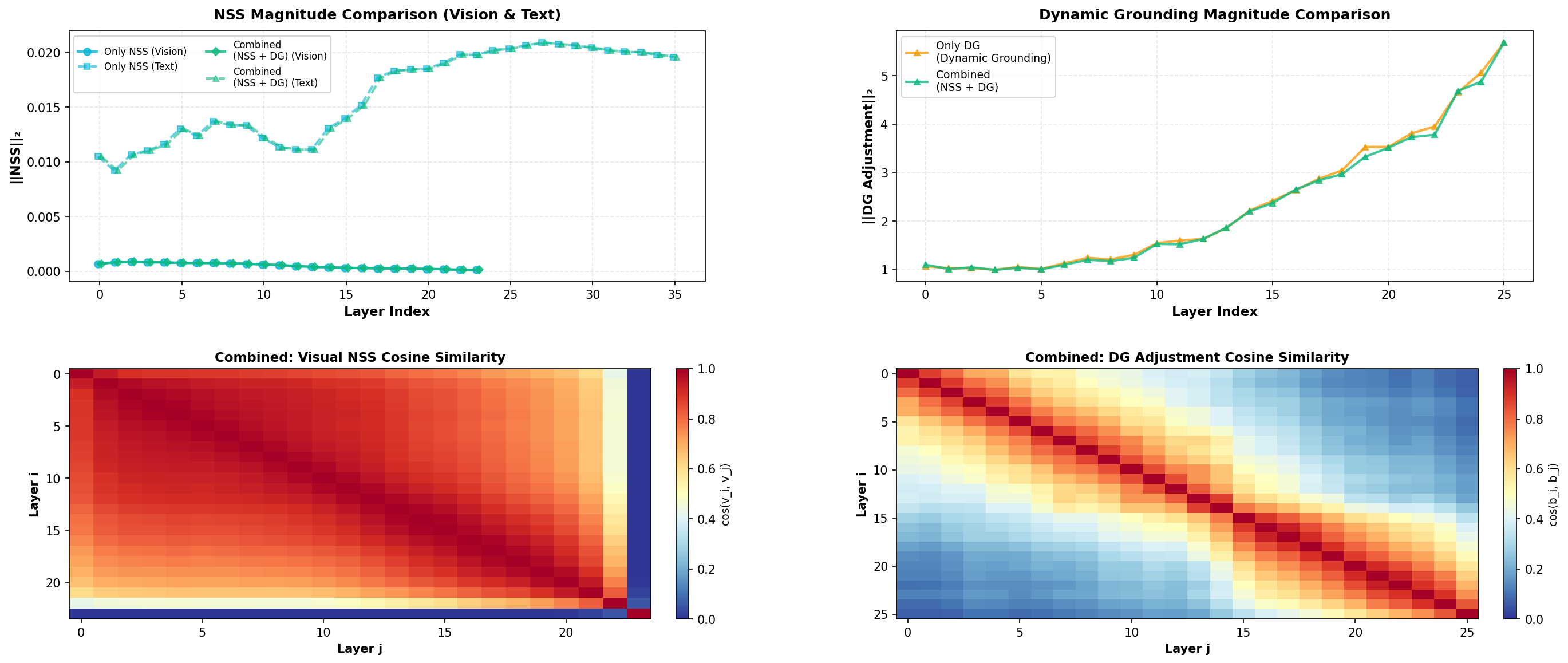}
   \caption{\textbf{Layer-wise magnitude and cross-layer similarity of Stage~1 (NSS) and Stage~2 (DG) updates.}
    Top: $\ell_2$ norms of NSS steering directions (vision/text) and DG adjustment vectors across layers, comparing stage-wise ablations (\textbf{Only NSS}, \textbf{Only DG}) with the \textbf{Combined} method. 
    Bottom: cosine-similarity matrices across layers for the visual NSS directions and DG adjustments under the \textbf{Combined} setting.
    NSS directions remain stable and globally coherent across layers, while DG adjustments exhibit more localized, layer-dependent structure, supporting the role of NSS as a global static prior and DG as an instance-wise grounding correction.}

   \Description{Layer-wise analysis of the intervention. The top panels plot the L2 magnitudes of NSS steering directions and DG adjustment vectors across layers. The bottom panels show cross-layer cosine-similarity matrices for the visual NSS directions and DG adjustments.}

   \label{fig:method_vis}
\end{figure*}

Under \textbf{Faithful} auxiliary text, we observe a largely conservative behavior: the combined intervention typically preserves the original performance and yields only small fluctuations around the frozen baseline. This is expected, since when the auxiliary text is already well-grounded and aligned with the visual evidence, the model’s hidden states are close to a ``consistent'' trajectory, leaving limited room for corrective steering. Nevertheless, a small number of model--dataset pairs exhibit noticeably negative deltas (e.g., $-3.1\%$, $-2.2\%$, $-2.1\%$). We attribute these drops to an inherent trade-off of robustness-oriented steering: Stage~2 (DG) explicitly down-weights text representations when it detects a mismatch between the auxiliary-text summary token and question-relevant visual evidence, and Stage~1 (NSS) further imposes a global stability prior that can slightly dampen beneficial text signals. In rare cases where the model already exploits Faithful text extremely effectively (or when the question can be answered with minimal visual grounding), these conservative shifts may mildly reduce the model’s reliance on auxiliary cues and thus lead to a modest accuracy decrease.

Importantly, such degradations remain limited in scope and do not change the overall conclusion of the combined method. The same hyperparameters that cause these rare Faithful regressions deliver consistent and often substantial gains in the \emph{harmful} regimes, where bias is most severe and where robustness is the primary objective. In particular, the method recovers large amounts of accuracy under \textbf{Option-Redirecting} auxiliary text (e.g., up to $+8.1\%$, $+5.1\%$, and $+6.6\%$ on VCR and Causal-VidQA), indicating that the intervention effectively suppresses option-targeted text bias without broadly collapsing the model’s ability to benefit from helpful descriptions. Therefore, even when a few Faithful cases incur moderate negative deltas, the net effect across ladder conditions demonstrates that the method maintains strong clean performance while substantially improving worst-case robustness against misleading, option-targeted auxiliary text.

Outside Option-Redirecting, performance changes are generally small.
Under \textbf{No-Aux}, \textbf{Irrelevant}, \textbf{Off-Question}, and \textbf{Contradictory} conditions,
the improvements typically remain within about $\pm 1\%$ (Figure~\ref{fig:method}), indicating that the combined intervention does not
introduce broad distributional shifts when the auxiliary text is absent or weakly informative.
This stability is important in practice, since real-world pipelines may provide auxiliary text of varying quality.

A remaining limitation is that \textbf{Faithful} text can occasionally see small-to-moderate drops (notably on VCR for some models).
This reflects the inherent tension faced by any bias-mitigation method: mechanisms that discount misleading auxiliary text can also
down-weight helpful text in some instances.
Nevertheless, combining NSS with DG is beneficial overall: compared to using DG alone, NSS acts as a conservative backbone that keeps representations more stable across conditions, while DG supplies the strong, instance-aware correction needed specifically for Option-Redirecting attacks. 
As a result, the full two-stage method achieves the best robustness profile: \emph{large targeted gains} where text bias is most harmful, without sacrificing stability across the remaining ladder conditions.

\paragraph{Limited impact when text is weakly informative.}
Under Off-Question and No-Aux, the average effect is near zero, which is expected: when auxiliary text is irrelevant to the
decision (or absent), there is less text-induced bias to correct, and a global shift should not materially change predictions.

Despite the above gains, NSS does not reliably recover from severe option-targeted collapse in all settings.
In particular, improvements on VCR under Option-Redirecting are less consistent and can be negative for some models,
highlighting a core constraint of Stage~1: NSS applies the same correction regardless of whether the current auxiliary text is Faithful,
Contradictory, or adversarially Option-Redirecting.
This motivates Stage~2 (DG), which is explicitly instance- and question-aware and can selectively realign auxiliary-text representations
with question-relevant visual evidence.

Overall, this ablation supports the intended division of labor: NSS provides a lightweight robustness baseline that stabilizes behavior across
conditions with essentially zero test-time overhead, while the substantial and reliably targeted mitigation of option-targeted failures
requires the dynamic grounding correction in Stage~2.

\subsubsection{Option-Redirecting (Worst-Case Text Bias)}
\label{app:exp-option-redirecting}

We highlight the \textbf{Option-Redirecting} condition, where the auxiliary text is question-aligned but steers the model toward the same incorrect option, causing the most severe degradation. As summarized in the combined results
(in Figure~\ref{fig:method}), our two-stage intervention yields the \emph{largest} accuracy recovery in this regime
(e.g., up to +14.6\% on ScienceQA, and consistently positive gains on A-OKVQA/VCR/Causal-VidQA), indicating that the
method effectively restores correct, visually grounded choices when text bias is strongest.

In the \textbf{Option-Redirecting} setting, each auxiliary text is constructed to subtly favor a specific \emph{target} incorrect option (the \emph{redirected option}).
We define \emph{Redirect} as the percentage of samples (out of 1{,}000) whose prediction matches the redirected option; lower values indicate stronger robustness to option-targeted textual bias.

Table~\ref{tab:redirect-analysis} reports this redirect count and confirms that the gains under Option-Redirecting are not merely aggregate accuracy fluctuations.
Across all models and datasets, our two-stage method consistently reduces the number of redirected predictions, indicating that it weakens the attraction to the biased option and shifts decisions back toward visually supported answers.
The reduction magnitude varies by dataset and model scale (typically larger on ScienceQA/A-OKVQA than on Causal-VidQA), but the trend is uniform: fewer redirected errors under the most adversarial auxiliary-text condition.

\subsubsection{Why Combining NSS and DG Is Necessary}
\label{app:why_combine}

Stage-wise ablations (Figures~\ref{fig:method_only_NSS}--\ref{fig:method_only_DG}) highlight a clear division of labor between the two stages.
When enabling \textbf{only Stage~1 (NSS)}, we observe modest but consistent gains across auxiliary-text conditions, with the key effect being \emph{stability}: performance does not degrade under misleading text and does not drop under Faithful descriptions. This aligns with NSS as a global, static correction that reduces sensitivity to superficial textual noise rather than resolving fine-grained grounding conflicts.

In contrast, \textbf{only Stage~2 (DG)} produces strong but selective improvements, concentrated under \textbf{Option-Redirecting} auxiliary text while remaining largely unchanged elsewhere. DG is instance-specific and activates precisely when the auxiliary text is question-aligned yet visually inconsistent and implicitly favors a particular option, in which case it realigns representations toward question-relevant visual evidence.
When auxiliary text is Faithful or weakly informative, the correction becomes negligible.

These results show that the stages are not interchangeable. NSS provides a robustness baseline that stabilizes representations against noisy textual cues, but is intentionally option-agnostic and therefore cannot fully correct option-targeted misdirection. DG corrects decision-level option bias, but benefits from operating on more stable upstream representations. Combining NSS and DG yields a balanced intervention:
NSS improves global stability, and DG provides fine-grained, instance-level correction precisely where harmful auxiliary text exerts the strongest influence.

Figure~\ref{fig:method_vis} further corroborates this division of labor at the representation level. 
The NSS vector norms (both vision and text) are almost identical between \textbf{Only NSS} and \textbf{Combined} across layers, indicating that Stage~1 provides a fixed, global steering prior that is estimated once and reused unchanged. 
In contrast, the DG adjustment magnitudes closely match between \textbf{Only DG} and \textbf{Combined} and increase with depth, showing that Stage~2 contributes an additional, sample-dependent correction on top of NSS rather than modifying it. 
Consistently, the cosine-similarity maps show broadly coherent NSS directions across layers, whereas DG adjustments exhibit a more localized (near-diagonal) similarity pattern, reflecting layer-wise, instance-specific grounding updates driven by question-relevant visual evidence.

Together, these trends explain why the combined method retains NSS’s global stability while enabling DG’s targeted correction precisely under Option-Redirecting text.

\subsection{Qualitative Examples under Option-Redirecting}
\label{app:example_visualize_method}
\begin{figure}
    \centering
    \includegraphics[width=1\linewidth]{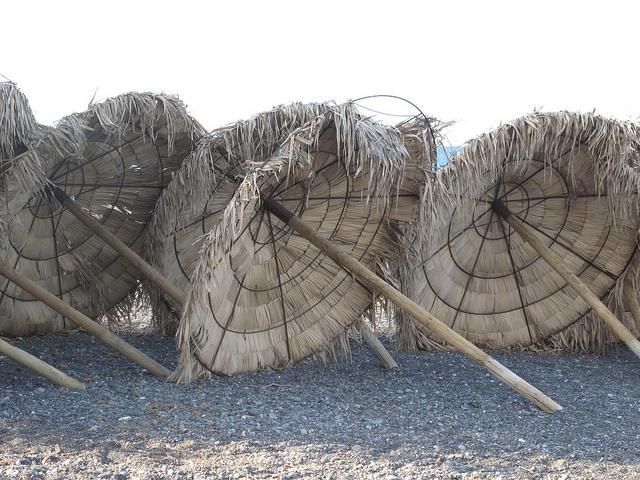}
    \caption{Input image for a successful correction case under Option-Redirecting auxiliary text.}
    \Description{Beach umbrellas viewed from underneath, showing circular thatched canopies supported by wooden poles.}
    \label{fig:train_img_00001821}
\end{figure}

\begin{figure}
    \centering
    \includegraphics[width=1\linewidth]{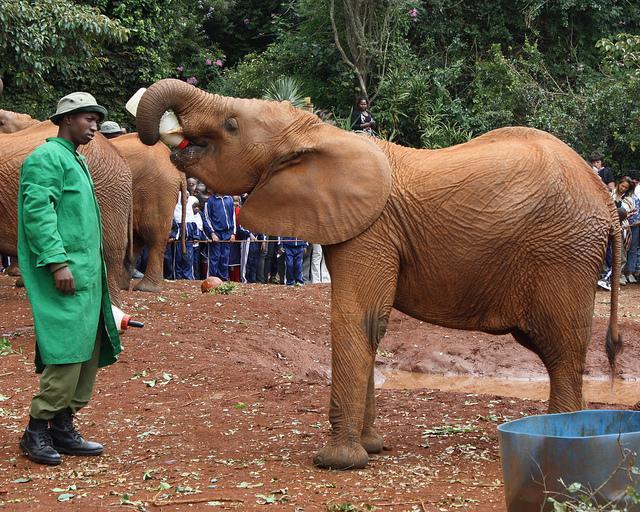}
    \caption{Input image for a failure case under Option-Redirecting auxiliary text, where the redirected distractor remains dominant after applying NSS+DG.}
    \Description{An elephant standing beside a keeper in a green coat, with the elephant raising its trunk near another elephant and a crowd in the background.}
    \label{fig:train_img_00008406}
\end{figure}

\begin{figure}
    \centering
    \includegraphics[width=1\linewidth]{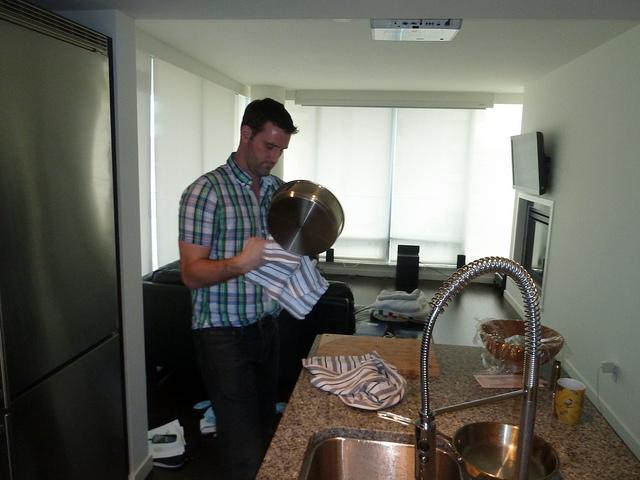}
    \caption{Input image for a successful correction case under Option-Redirecting auxiliary text.}
    \Description{A man in a kitchen drying a metal pot with a striped towel beside a sink.}
    \label{fig:train_img_00007245}
\end{figure}

\begin{figure}
    \centering
    \includegraphics[width=1\linewidth]{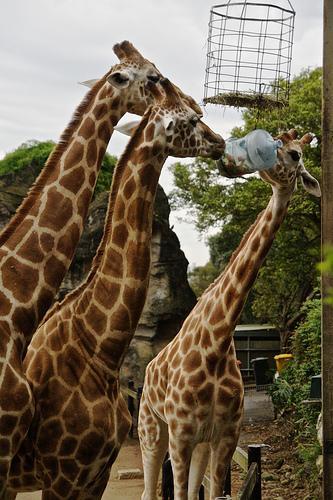}
    \caption{Input image for a failure case under Option-Redirecting auxiliary text.}
    \Description{Three giraffes reaching toward a suspended feeder, with one giraffe extending its neck upward.}
    \label{fig:train_img_00007994}
\end{figure}

\begin{figure}
    \centering
    \includegraphics[width=1\linewidth]{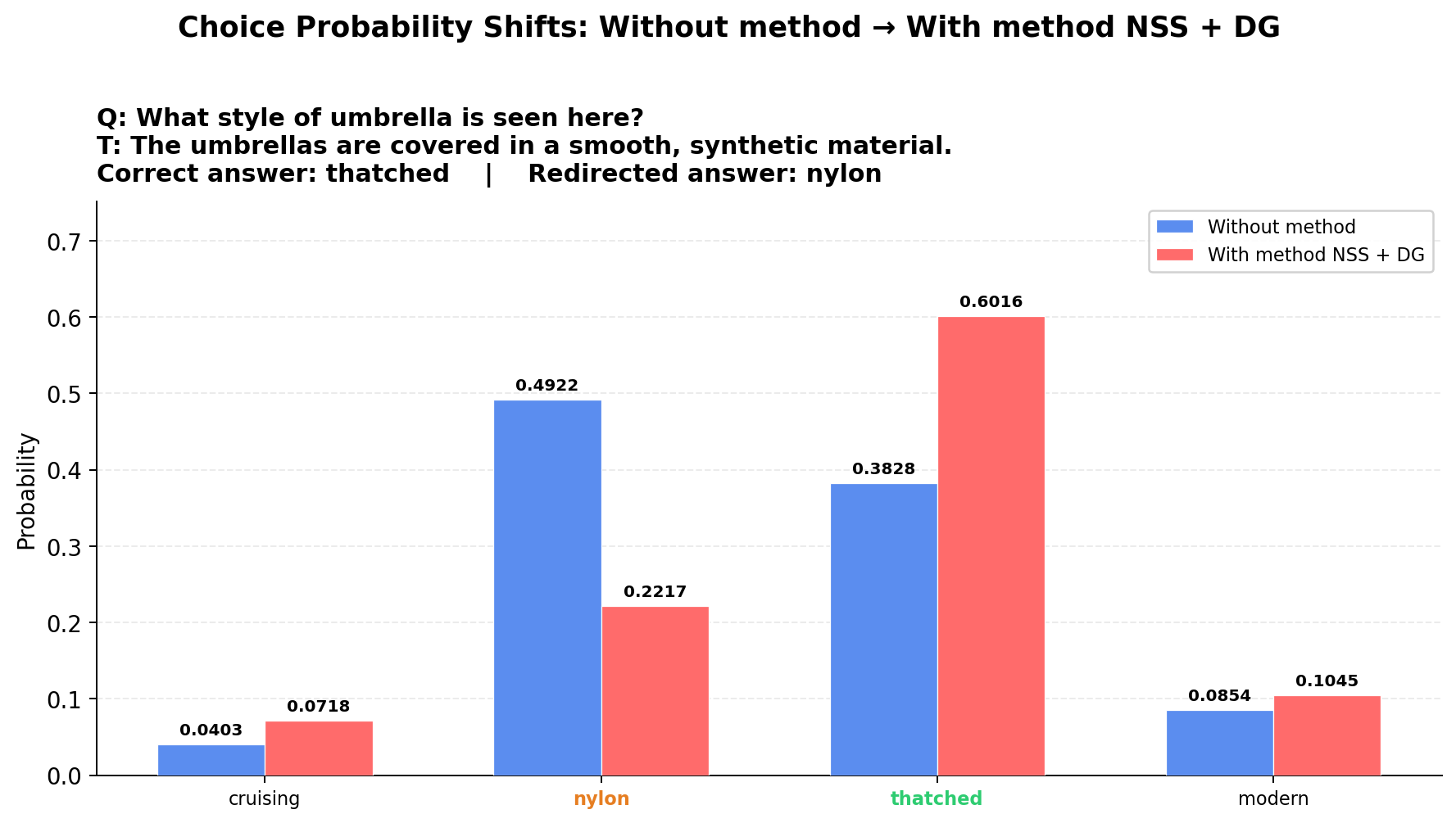}
    \caption{Choice-level probability shift for the example in Figure~\ref{fig:train_img_00001821}. NSS+DG reduces the redirected distractor and increases the correct answer probability.}
    \Description{Grouped bar chart of answer-choice probabilities before and after NSS plus DG for the umbrella example. Probability decreases for the redirected answer nylon and increases for the correct answer thatched.}
    \label{fig:logit_train_img_00001821}
\end{figure}

\begin{figure}
    \centering
    \includegraphics[width=1\linewidth]{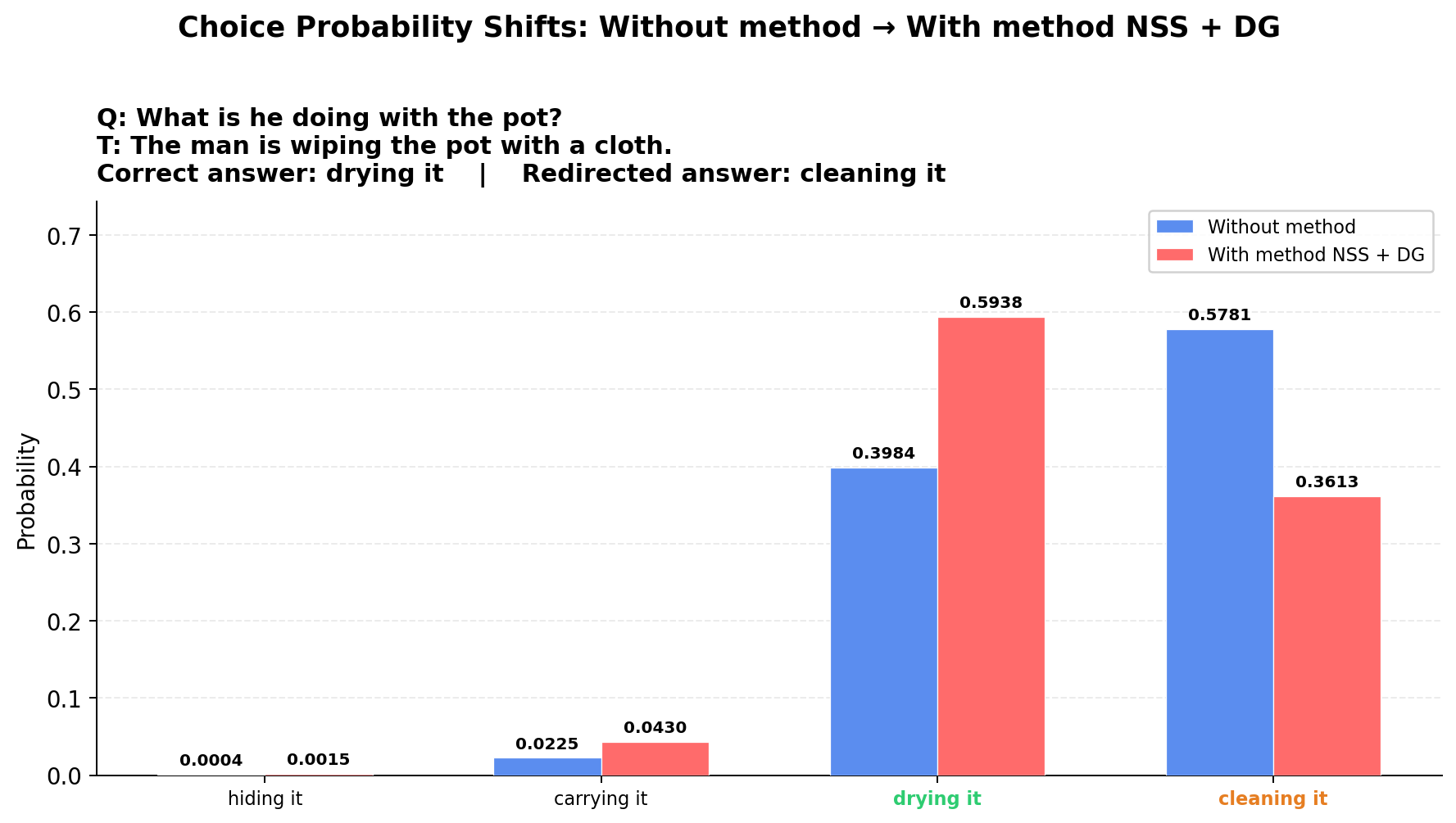}
    \caption{Choice-level probability shift for the example in Figure~\ref{fig:train_img_00007245}. The intervention successfully rebalances probability mass toward the visually grounded answer.}
    \Description{Grouped bar chart of answer-choice probabilities before and after NSS plus DG for the kitchen example, showing probability mass shifting toward the visually grounded correct answer.}
    \label{fig:logit_train_img_00007245}
\end{figure}

\begin{figure}
    \centering
    \includegraphics[width=1\linewidth]{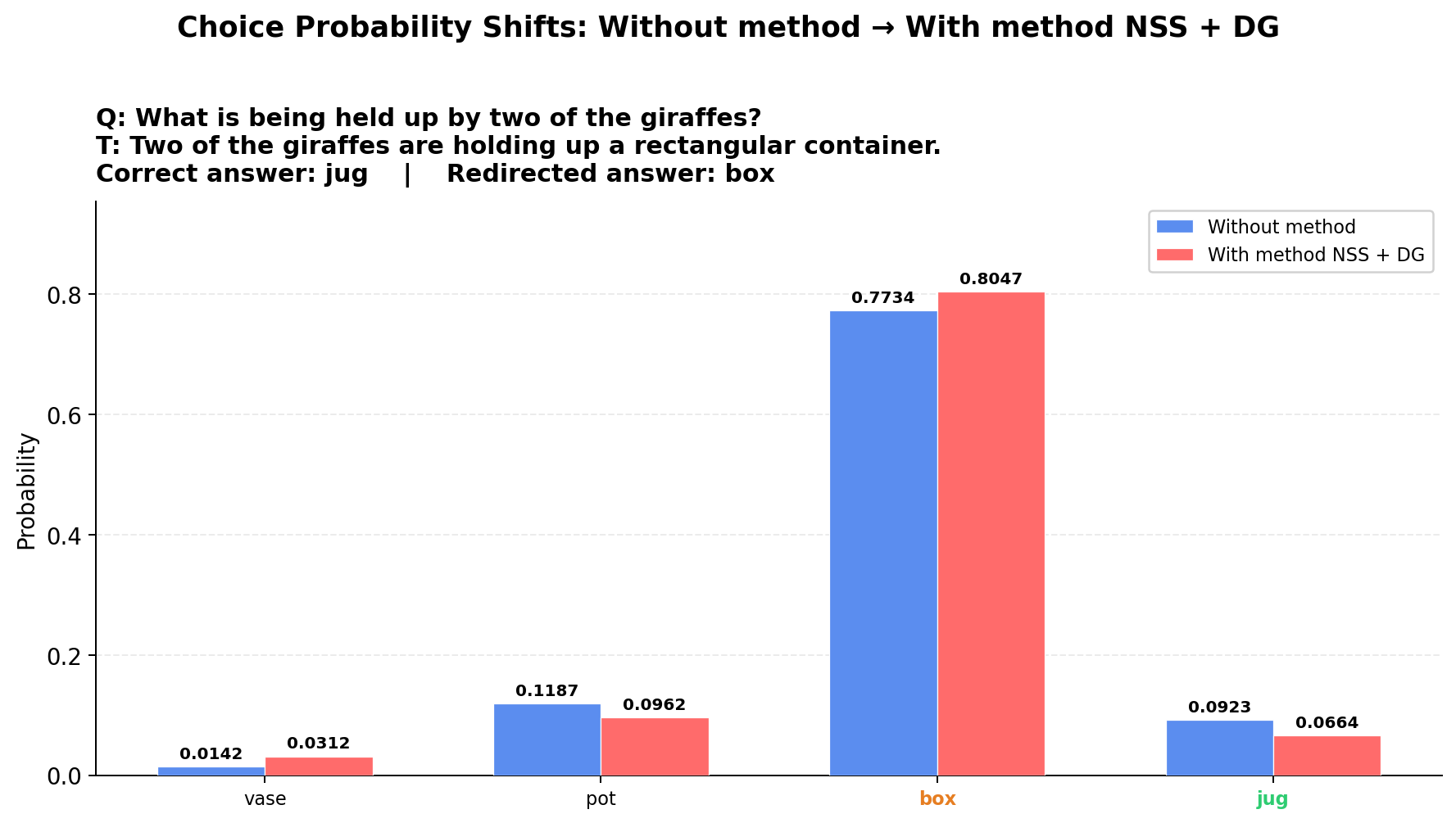}
    \caption{Choice-level probability shift for the example in Figure~\ref{fig:train_img_00007994}. Although NSS+DG changes the score distribution, the redirected distractor remains competitive, resulting in failure.}
    \Description{Grouped bar chart of answer-choice probabilities before and after NSS plus DG for the giraffe example. The score distribution changes but the redirected distractor remains competitive.}
    \label{fig:logit_train_img_00007994}
\end{figure}

\begin{figure}
    \centering
    \includegraphics[width=1\linewidth]{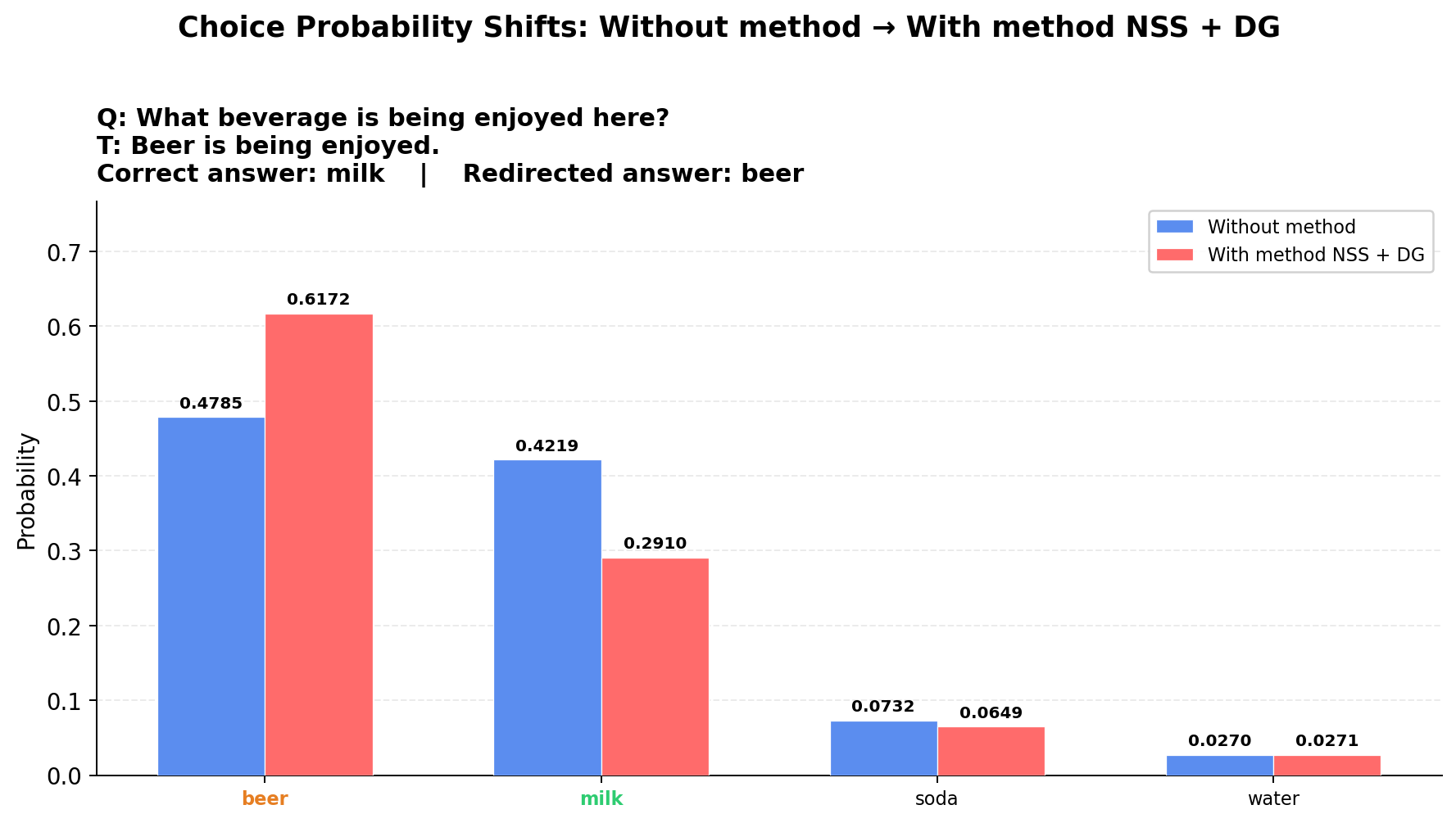}
    \caption{Choice-level probability shift for the example in Figure~\ref{fig:train_img_00008406}. The redirected distractor remains dominant after intervention, illustrating a failure case.}
    \Description{Grouped bar chart of answer-choice probabilities before and after NSS plus DG for the elephant example. The redirected distractor remains dominant after the intervention.}
    \label{fig:logit_train_img_00008406}
\end{figure}

To provide a finer-grained view beyond dataset-level averages, we present qualitative examples under the \textbf{Option-Redirecting} setting and visualize the corresponding choice probabilities before and after applying \textbf{NSS+DG} (Figures~\ref{fig:train_img_00001821}--\ref{fig:logit_train_img_00008406}). These cases directly illustrate the decision-level phenomenon studied in this work: misleading yet question-relevant auxiliary text can concentrate probability mass on a specific redirected distractor, despite conflicting visual evidence. Successful examples (Figures~\ref{fig:train_img_00001821}, \ref{fig:train_img_00007245}, \ref{fig:logit_train_img_00001821}, and \ref{fig:logit_train_img_00007245}) show that the proposed intervention can reduce this redirected preference and restore probability mass to the visually supported correct answer. Failure examples (Figures~\ref{fig:train_img_00007994}, \ref{fig:train_img_00008406}, \ref{fig:logit_train_img_00007994}, and \ref{fig:logit_train_img_00008406}), in contrast, indicate that such correction is not always sufficient when the option-targeted textual signal is particularly strong. Together, these examples offer an instance-level interpretation of the robustness gains and residual errors observed in the quantitative analysis.


\begin{thebibliography}{43}


\ifx \showCODEN    \undefined \def \showCODEN     #1{\unskip}     \fi
\ifx \showISBNx    \undefined \def \showISBNx     #1{\unskip}     \fi
\ifx \showISBNxiii \undefined \def \showISBNxiii  #1{\unskip}     \fi
\ifx \showISSN     \undefined \def \showISSN      #1{\unskip}     \fi
\ifx \showLCCN     \undefined \def \showLCCN      #1{\unskip}     \fi
\ifx \shownote     \undefined \def \shownote      #1{#1}          \fi
\ifx \showarticletitle \undefined \def \showarticletitle #1{#1}   \fi
\ifx \showURL      \undefined \def \showURL       {\relax}        \fi
\providecommand\bibfield[2]{#2}
\providecommand\bibinfo[2]{#2}
\providecommand\natexlab[1]{#1}
\providecommand\showeprint[2][]{arXiv:#2}

\bibitem[Agrawal et~al\mbox{.}(2018)]%
        {agrawal2018don}
\bibfield{author}{\bibinfo{person}{Aishwarya Agrawal}, \bibinfo{person}{Dhruv
  Batra}, \bibinfo{person}{Devi Parikh}, {and} \bibinfo{person}{Aniruddha
  Kembhavi}.} \bibinfo{year}{2018}\natexlab{}.
\newblock \showarticletitle{Don't just assume; look and answer: Overcoming
  priors for visual question answering}. In
  \bibinfo{booktitle}{\emph{Proceedings of the IEEE conference on computer
  vision and pattern recognition}}. \bibinfo{pages}{4971--4980}.
\newblock


\bibitem[Bai et~al\mbox{.}(2025a)]%
        {bai2025qwen3vltechnicalreport}
\bibfield{author}{\bibinfo{person}{Shuai Bai}, \bibinfo{person}{Yuxuan Cai},
  \bibinfo{person}{Ruizhe Chen}, \bibinfo{person}{Keqin Chen},
  \bibinfo{person}{Xionghui Chen}, \bibinfo{person}{Zesen Cheng},
  \bibinfo{person}{Lianghao Deng}, \bibinfo{person}{Wei Ding},
  \bibinfo{person}{Chang Gao}, \bibinfo{person}{Chunjiang Ge},
  \bibinfo{person}{Wenbin Ge}, \bibinfo{person}{Zhifang Guo},
  \bibinfo{person}{Qidong Huang}, \bibinfo{person}{Jie Huang},
  \bibinfo{person}{Fei Huang}, \bibinfo{person}{Binyuan Hui},
  \bibinfo{person}{Shutong Jiang}, \bibinfo{person}{Zhaohai Li},
  \bibinfo{person}{Mingsheng Li}, \bibinfo{person}{Mei Li},
  \bibinfo{person}{Kaixin Li}, \bibinfo{person}{Zicheng Lin},
  \bibinfo{person}{Junyang Lin}, \bibinfo{person}{Xuejing Liu},
  \bibinfo{person}{Jiawei Liu}, \bibinfo{person}{Chenglong Liu},
  \bibinfo{person}{Yang Liu}, \bibinfo{person}{Dayiheng Liu},
  \bibinfo{person}{Shixuan Liu}, \bibinfo{person}{Dunjie Lu},
  \bibinfo{person}{Ruilin Luo}, \bibinfo{person}{Chenxu Lv},
  \bibinfo{person}{Rui Men}, \bibinfo{person}{Lingchen Meng},
  \bibinfo{person}{Xuancheng Ren}, \bibinfo{person}{Xingzhang Ren},
  \bibinfo{person}{Sibo Song}, \bibinfo{person}{Yuchong Sun},
  \bibinfo{person}{Jun Tang}, \bibinfo{person}{Jianhong Tu},
  \bibinfo{person}{Jianqiang Wan}, \bibinfo{person}{Peng Wang},
  \bibinfo{person}{Pengfei Wang}, \bibinfo{person}{Qiuyue Wang},
  \bibinfo{person}{Yuxuan Wang}, \bibinfo{person}{Tianbao Xie},
  \bibinfo{person}{Yiheng Xu}, \bibinfo{person}{Haiyang Xu},
  \bibinfo{person}{Jin Xu}, \bibinfo{person}{Zhibo Yang},
  \bibinfo{person}{Mingkun Yang}, \bibinfo{person}{Jianxin Yang},
  \bibinfo{person}{An Yang}, \bibinfo{person}{Bowen Yu}, \bibinfo{person}{Fei
  Zhang}, \bibinfo{person}{Hang Zhang}, \bibinfo{person}{Xi Zhang},
  \bibinfo{person}{Bo Zheng}, \bibinfo{person}{Humen Zhong},
  \bibinfo{person}{Jingren Zhou}, \bibinfo{person}{Fan Zhou},
  \bibinfo{person}{Jing Zhou}, \bibinfo{person}{Yuanzhi Zhu}, {and}
  \bibinfo{person}{Ke Zhu}.} \bibinfo{year}{2025}\natexlab{a}.
\newblock \bibinfo{title}{Qwen3-VL Technical Report}.
\newblock
\showeprint[arxiv]{2511.21631}~[cs.CV]
\urldef\tempurl%
\url{https://arxiv.org/abs/2511.21631}
\showURL{%
\tempurl}


\bibitem[Bai et~al\mbox{.}(2025b)]%
        {bai2025qwen2}
\bibfield{author}{\bibinfo{person}{Shuai Bai}, \bibinfo{person}{Keqin Chen},
  \bibinfo{person}{Xuejing Liu}, \bibinfo{person}{Jialin Wang},
  \bibinfo{person}{Wenbin Ge}, \bibinfo{person}{Sibo Song},
  \bibinfo{person}{Kai Dang}, \bibinfo{person}{Peng Wang},
  \bibinfo{person}{Shijie Wang}, \bibinfo{person}{Jun Tang}, {et~al\mbox{.}}}
  \bibinfo{year}{2025}\natexlab{b}.
\newblock \showarticletitle{Qwen2. 5-vl technical report}.
\newblock \bibinfo{journal}{\emph{arXiv preprint arXiv:2502.13923}}
  (\bibinfo{year}{2025}).
\newblock


\bibitem[Bai et~al\mbox{.}(2024)]%
        {bai2024hallucination}
\bibfield{author}{\bibinfo{person}{Zechen Bai}, \bibinfo{person}{Pichao Wang},
  \bibinfo{person}{Tianjun Xiao}, \bibinfo{person}{Tong He},
  \bibinfo{person}{Zongbo Han}, \bibinfo{person}{Zheng Zhang}, {and}
  \bibinfo{person}{Mike~Zheng Shou}.} \bibinfo{year}{2024}\natexlab{}.
\newblock \showarticletitle{Hallucination of multimodal large language models:
  A survey}.
\newblock \bibinfo{journal}{\emph{arXiv preprint arXiv:2404.18930}}
  (\bibinfo{year}{2024}).
\newblock


\bibitem[Cadene et~al\mbox{.}(2019)]%
        {cadene2019rubi}
\bibfield{author}{\bibinfo{person}{Remi Cadene}, \bibinfo{person}{Corentin
  Dancette}, \bibinfo{person}{Matthieu Cord}, \bibinfo{person}{Devi Parikh},
  {et~al\mbox{.}}} \bibinfo{year}{2019}\natexlab{}.
\newblock \showarticletitle{Rubi: Reducing unimodal biases for visual question
  answering}.
\newblock \bibinfo{journal}{\emph{Advances in neural information processing
  systems}}  \bibinfo{volume}{32} (\bibinfo{year}{2019}).
\newblock


\bibitem[Chen et~al\mbox{.}(2025)]%
        {chen2025livecc}
\bibfield{author}{\bibinfo{person}{Joya Chen}, \bibinfo{person}{Ziyun Zeng},
  \bibinfo{person}{Yiqi Lin}, \bibinfo{person}{Wei Li}, \bibinfo{person}{Zejun
  Ma}, {and} \bibinfo{person}{Mike~Zheng Shou}.}
  \bibinfo{year}{2025}\natexlab{}.
\newblock \showarticletitle{Livecc: Learning video llm with streaming speech
  transcription at scale}. In \bibinfo{booktitle}{\emph{Proceedings of the
  Computer Vision and Pattern Recognition Conference}}.
  \bibinfo{pages}{29083--29095}.
\newblock


\bibitem[Comanici et~al\mbox{.}(2025)]%
        {comanici2025gemini}
\bibfield{author}{\bibinfo{person}{Gheorghe Comanici}, \bibinfo{person}{Eric
  Bieber}, \bibinfo{person}{Mike Schaekermann}, \bibinfo{person}{Ice Pasupat},
  \bibinfo{person}{Noveen Sachdeva}, \bibinfo{person}{Inderjit Dhillon},
  \bibinfo{person}{Marcel Blistein}, \bibinfo{person}{Ori Ram},
  \bibinfo{person}{Dan Zhang}, \bibinfo{person}{Evan Rosen}, {et~al\mbox{.}}}
  \bibinfo{year}{2025}\natexlab{}.
\newblock \showarticletitle{Gemini 2.5: Pushing the frontier with advanced
  reasoning, multimodality, long context, and next generation agentic
  capabilities}.
\newblock \bibinfo{journal}{\emph{arXiv preprint arXiv:2507.06261}}
  (\bibinfo{year}{2025}).
\newblock


\bibitem[Deng et~al\mbox{.}(2025)]%
        {deng2503words}
\bibfield{author}{\bibinfo{person}{Ailin Deng}, \bibinfo{person}{Tri Cao},
  \bibinfo{person}{Zhirui Chen}, {and} \bibinfo{person}{Bryan Hooi}.}
  \bibinfo{year}{2025}\natexlab{}.
\newblock \showarticletitle{Words or vision: Do vision-language models have
  blind faith in text?, 2025}.
\newblock \bibinfo{journal}{\emph{URL https://arxiv. org/abs/2503.02199}}
  (\bibinfo{year}{2025}).
\newblock


\bibitem[Fu et~al\mbox{.}(2024)]%
        {fu2024ocrbench}
\bibfield{author}{\bibinfo{person}{Ling Fu}, \bibinfo{person}{Zhebin Kuang},
  \bibinfo{person}{Jiajun Song}, \bibinfo{person}{Mingxin Huang},
  \bibinfo{person}{Biao Yang}, \bibinfo{person}{Yuzhe Li},
  \bibinfo{person}{Linghao Zhu}, \bibinfo{person}{Qidi Luo},
  \bibinfo{person}{Xinyu Wang}, \bibinfo{person}{Hao Lu}, {et~al\mbox{.}}}
  \bibinfo{year}{2024}\natexlab{}.
\newblock \showarticletitle{Ocrbench v2: An improved benchmark for evaluating
  large multimodal models on visual text localization and reasoning}.
\newblock \bibinfo{journal}{\emph{arXiv preprint arXiv:2501.00321}}
  (\bibinfo{year}{2024}).
\newblock


\bibitem[Fu et~al\mbox{.}(2025)]%
        {fu2025hidden}
\bibfield{author}{\bibinfo{person}{Stephanie Fu}, \bibinfo{person}{Tyler
  Bonnen}, \bibinfo{person}{Devin Guillory}, {and} \bibinfo{person}{Trevor
  Darrell}.} \bibinfo{year}{2025}\natexlab{}.
\newblock \showarticletitle{Hidden in plain sight: VLMs overlook their visual
  representations}.
\newblock \bibinfo{journal}{\emph{arXiv preprint arXiv:2506.08008}}
  (\bibinfo{year}{2025}).
\newblock


\bibitem[Gan et~al\mbox{.}(2025)]%
        {gan2025textual}
\bibfield{author}{\bibinfo{person}{Woody~Haosheng Gan}, \bibinfo{person}{Deqing
  Fu}, \bibinfo{person}{Julian Asilis}, \bibinfo{person}{Ollie Liu},
  \bibinfo{person}{Dani Yogatama}, \bibinfo{person}{Vatsal Sharan},
  \bibinfo{person}{Robin Jia}, {and} \bibinfo{person}{Willie Neiswanger}.}
  \bibinfo{year}{2025}\natexlab{}.
\newblock \showarticletitle{Textual Steering Vectors Can Improve Visual
  Understanding in Multimodal Large Language Models}.
\newblock \bibinfo{journal}{\emph{arXiv preprint arXiv:2505.14071}}
  (\bibinfo{year}{2025}).
\newblock


\bibitem[Guo et~al\mbox{.}(2021)]%
        {guo2021loss}
\bibfield{author}{\bibinfo{person}{Yangyang Guo}, \bibinfo{person}{Liqiang
  Nie}, \bibinfo{person}{Zhiyong Cheng}, \bibinfo{person}{Qi Tian}, {and}
  \bibinfo{person}{Min Zhang}.} \bibinfo{year}{2021}\natexlab{}.
\newblock \showarticletitle{Loss re-scaling VQA: Revisiting the language prior
  problem from a class-imbalance view}.
\newblock \bibinfo{journal}{\emph{IEEE Transactions on Image Processing}}
  \bibinfo{volume}{31} (\bibinfo{year}{2021}), \bibinfo{pages}{227--238}.
\newblock


\bibitem[Hua et~al\mbox{.}(2025)]%
        {hua2025vision}
\bibfield{author}{\bibinfo{person}{Tianze Hua}, \bibinfo{person}{Tian Yun},
  {and} \bibinfo{person}{Ellie Pavlick}.} \bibinfo{year}{2025}\natexlab{}.
\newblock \showarticletitle{How Do Vision-Language Models Process Conflicting
  Information Across Modalities?}
\newblock \bibinfo{journal}{\emph{arXiv preprint arXiv:2507.01790}}
  (\bibinfo{year}{2025}).
\newblock


\bibitem[Huang et~al\mbox{.}(2024)]%
        {huang2024opera}
\bibfield{author}{\bibinfo{person}{Qidong Huang}, \bibinfo{person}{Xiaoyi
  Dong}, \bibinfo{person}{Pan Zhang}, \bibinfo{person}{Bin Wang},
  \bibinfo{person}{Conghui He}, \bibinfo{person}{Jiaqi Wang},
  \bibinfo{person}{Dahua Lin}, \bibinfo{person}{Weiming Zhang}, {and}
  \bibinfo{person}{Nenghai Yu}.} \bibinfo{year}{2024}\natexlab{}.
\newblock \showarticletitle{Opera: Alleviating hallucination in multi-modal
  large language models via over-trust penalty and retrospection-allocation}.
  In \bibinfo{booktitle}{\emph{Proceedings of the IEEE/CVF Conference on
  Computer Vision and Pattern Recognition}}. \bibinfo{pages}{13418--13427}.
\newblock


\bibitem[Jian et~al\mbox{.}(2024)]%
        {jian2024large}
\bibfield{author}{\bibinfo{person}{Pu Jian}, \bibinfo{person}{Donglei Yu},
  {and} \bibinfo{person}{Jiajun Zhang}.} \bibinfo{year}{2024}\natexlab{}.
\newblock \showarticletitle{Large language models know what is key visual
  entity: An LLM-assisted multimodal retrieval for VQA}. In
  \bibinfo{booktitle}{\emph{Proceedings of the 2024 Conference on Empirical
  Methods in Natural Language Processing}}. \bibinfo{pages}{10939--10956}.
\newblock


\bibitem[Jiang et~al\mbox{.}(2025)]%
        {jiang2025memory}
\bibfield{author}{\bibinfo{person}{Hongda Jiang}, \bibinfo{person}{Xinyuan
  Zhang}, \bibinfo{person}{Siddhant Garg}, \bibinfo{person}{Rishab Arora},
  \bibinfo{person}{Shiun-Zu Kuo}, \bibinfo{person}{Jiayang Xu},
  \bibinfo{person}{Aaron Colak}, {and} \bibinfo{person}{Xin~Luna Dong}.}
  \bibinfo{year}{2025}\natexlab{}.
\newblock \showarticletitle{Memory-QA: Answering Recall Questions Based on
  Multimodal Memories}. In \bibinfo{booktitle}{\emph{Proceedings of the 2025
  Conference on Empirical Methods in Natural Language Processing}}.
  \bibinfo{pages}{24255--24277}.
\newblock


\bibitem[Leng et~al\mbox{.}(2024)]%
        {leng2024mitigating}
\bibfield{author}{\bibinfo{person}{Sicong Leng}, \bibinfo{person}{Hang Zhang},
  \bibinfo{person}{Guanzheng Chen}, \bibinfo{person}{Xin Li},
  \bibinfo{person}{Shijian Lu}, \bibinfo{person}{Chunyan Miao}, {and}
  \bibinfo{person}{Lidong Bing}.} \bibinfo{year}{2024}\natexlab{}.
\newblock \showarticletitle{Mitigating object hallucinations in large
  vision-language models through visual contrastive decoding}. In
  \bibinfo{booktitle}{\emph{Proceedings of the IEEE/CVF Conference on Computer
  Vision and Pattern Recognition}}. \bibinfo{pages}{13872--13882}.
\newblock


\bibitem[Li et~al\mbox{.}(2022)]%
        {li2022representation}
\bibfield{author}{\bibinfo{person}{Jiangtong Li}, \bibinfo{person}{Li Niu},
  {and} \bibinfo{person}{Liqing Zhang}.} \bibinfo{year}{2022}\natexlab{}.
\newblock \showarticletitle{From representation to reasoning: Towards both
  evidence and commonsense reasoning for video question-answering}. In
  \bibinfo{booktitle}{\emph{Proceedings of the IEEE/CVF conference on computer
  vision and pattern recognition}}. \bibinfo{pages}{21273--21282}.
\newblock


\bibitem[Lin et~al\mbox{.}(2023)]%
        {lin2023revisiting}
\bibfield{author}{\bibinfo{person}{Zhiqiu Lin}, \bibinfo{person}{Xinyue Chen},
  \bibinfo{person}{Deepak Pathak}, \bibinfo{person}{Pengchuan Zhang}, {and}
  \bibinfo{person}{Deva Ramanan}.} \bibinfo{year}{2023}\natexlab{}.
\newblock \showarticletitle{Revisiting the role of language priors in
  vision-language models}.
\newblock \bibinfo{journal}{\emph{arXiv preprint arXiv:2306.01879}}
  (\bibinfo{year}{2023}).
\newblock


\bibitem[Liu et~al\mbox{.}(2024b)]%
        {liu2024survey}
\bibfield{author}{\bibinfo{person}{Hanchao Liu}, \bibinfo{person}{Wenyuan Xue},
  \bibinfo{person}{Yifei Chen}, \bibinfo{person}{Dapeng Chen},
  \bibinfo{person}{Xiutian Zhao}, \bibinfo{person}{Ke Wang},
  \bibinfo{person}{Liping Hou}, \bibinfo{person}{Rongjun Li}, {and}
  \bibinfo{person}{Wei Peng}.} \bibinfo{year}{2024}\natexlab{b}.
\newblock \showarticletitle{A survey on hallucination in large vision-language
  models}.
\newblock \bibinfo{journal}{\emph{arXiv preprint arXiv:2402.00253}}
  (\bibinfo{year}{2024}).
\newblock


\bibitem[Liu et~al\mbox{.}(2025)]%
        {liu2025reducing}
\bibfield{author}{\bibinfo{person}{Sheng Liu}, \bibinfo{person}{Haotian Ye},
  {and} \bibinfo{person}{James Zou}.} \bibinfo{year}{2025}\natexlab{}.
\newblock \showarticletitle{Reducing hallucinations in large vision-language
  models via latent space steering}. In \bibinfo{booktitle}{\emph{The
  Thirteenth International Conference on Learning Representations}}.
\newblock


\bibitem[Liu et~al\mbox{.}(2024a)]%
        {liu2024insight}
\bibfield{author}{\bibinfo{person}{Xiaoyuan Liu}, \bibinfo{person}{Wenxuan
  Wang}, \bibinfo{person}{Youliang Yuan}, \bibinfo{person}{Jen-tse Huang},
  \bibinfo{person}{Qiuzhi Liu}, \bibinfo{person}{Pinjia He}, {and}
  \bibinfo{person}{Zhaopeng Tu}.} \bibinfo{year}{2024}\natexlab{a}.
\newblock \showarticletitle{Insight Over Sight: Exploring the Vision-Knowledge
  Conflicts in Multimodal LLMs}.
\newblock \bibinfo{journal}{\emph{arXiv preprint arXiv:2410.08145}}
  (\bibinfo{year}{2024}).
\newblock


\bibitem[Lu et~al\mbox{.}(2022)]%
        {lu2022learn}
\bibfield{author}{\bibinfo{person}{Pan Lu}, \bibinfo{person}{Swaroop Mishra},
  \bibinfo{person}{Tanglin Xia}, \bibinfo{person}{Liang Qiu},
  \bibinfo{person}{Kai-Wei Chang}, \bibinfo{person}{Song-Chun Zhu},
  \bibinfo{person}{Oyvind Tafjord}, \bibinfo{person}{Peter Clark}, {and}
  \bibinfo{person}{Ashwin Kalyan}.} \bibinfo{year}{2022}\natexlab{}.
\newblock \showarticletitle{Learn to explain: Multimodal reasoning via thought
  chains for science question answering}.
\newblock \bibinfo{journal}{\emph{Advances in Neural Information Processing
  Systems}}  \bibinfo{volume}{35} (\bibinfo{year}{2022}),
  \bibinfo{pages}{2507--2521}.
\newblock


\bibitem[Merity et~al\mbox{.}(2016)]%
        {merity2016pointer}
\bibfield{author}{\bibinfo{person}{Stephen Merity}, \bibinfo{person}{Caiming
  Xiong}, \bibinfo{person}{James Bradbury}, {and} \bibinfo{person}{Richard
  Socher}.} \bibinfo{year}{2016}\natexlab{}.
\newblock \showarticletitle{Pointer sentinel mixture models}.
\newblock \bibinfo{journal}{\emph{arXiv preprint arXiv:1609.07843}}
  (\bibinfo{year}{2016}).
\newblock


\bibitem[Nguyen et~al\mbox{.}(2025)]%
        {nguyen2025challenges}
\bibfield{author}{\bibinfo{person}{Trang Nguyen}, \bibinfo{person}{Jackson
  Michaels}, \bibinfo{person}{Madalina Fiterau}, {and} \bibinfo{person}{David
  Jensen}.} \bibinfo{year}{2025}\natexlab{}.
\newblock \showarticletitle{Challenges in Understanding Modality Conflict in
  Vision-Language Models}.
\newblock \bibinfo{journal}{\emph{arXiv preprint arXiv:2509.02805}}
  (\bibinfo{year}{2025}).
\newblock


\bibitem[Park et~al\mbox{.}(2025)]%
        {park2025steer}
\bibfield{author}{\bibinfo{person}{Seongheon Park}, \bibinfo{person}{Xuefeng
  Du}, \bibinfo{person}{Min-Hsuan Yeh}, \bibinfo{person}{Haobo Wang}, {and}
  \bibinfo{person}{Yixuan Li}.} \bibinfo{year}{2025}\natexlab{}.
\newblock \showarticletitle{Steer LLM Latents for Hallucination Detection}.
\newblock \bibinfo{journal}{\emph{arXiv preprint arXiv:2503.01917}}
  (\bibinfo{year}{2025}).
\newblock


\bibitem[Popordanoska et~al\mbox{.}(2025)]%
        {popordanoskacrosscheck}
\bibfield{author}{\bibinfo{person}{Teodora Popordanoska},
  \bibinfo{person}{Jiameng Li}, {and} \bibinfo{person}{Matthew~B Blaschko}.}
  \bibinfo{year}{2025}\natexlab{}.
\newblock \showarticletitle{CrossCheck: A Vision-Language Conflict Detection
  Benchmark}.
\newblock  (\bibinfo{year}{2025}).
\newblock


\bibitem[Qraitem et~al\mbox{.}(2025)]%
        {qraitem2025web}
\bibfield{author}{\bibinfo{person}{Maan Qraitem}, \bibinfo{person}{Piotr
  Teterwak}, \bibinfo{person}{Kate Saenko}, {and} \bibinfo{person}{Bryan~A
  Plummer}.} \bibinfo{year}{2025}\natexlab{}.
\newblock \showarticletitle{Web Artifact Attacks Disrupt Vision Language
  Models}.
\newblock \bibinfo{journal}{\emph{arXiv preprint arXiv:2503.13652}}
  (\bibinfo{year}{2025}).
\newblock


\bibitem[Schwenk et~al\mbox{.}(2022)]%
        {schwenk2022okvqa}
\bibfield{author}{\bibinfo{person}{Dustin Schwenk}, \bibinfo{person}{Apoorv
  Khandelwal}, \bibinfo{person}{Christopher Clark}, \bibinfo{person}{Kenneth
  Marino}, {and} \bibinfo{person}{Roozbeh Mottaghi}.}
  \bibinfo{year}{2022}\natexlab{}.
\newblock \showarticletitle{A-okvqa: A benchmark for visual question answering
  using world knowledge}. In \bibinfo{booktitle}{\emph{European conference on
  computer vision}}. Springer, \bibinfo{pages}{146--162}.
\newblock


\bibitem[Sharifdeen et~al\mbox{.}(2025)]%
        {sharifdeen2025tpt}
\bibfield{author}{\bibinfo{person}{Ashshak Sharifdeen},
  \bibinfo{person}{Muhammad~Akhtar Munir}, \bibinfo{person}{Sanoojan Baliah},
  \bibinfo{person}{Salman Khan}, {and} \bibinfo{person}{Muhammad~Haris Khan}.}
  \bibinfo{year}{2025}\natexlab{}.
\newblock \showarticletitle{O-TPT: Orthogonality Constraints for Calibrating
  Test-time Prompt Tuning in Vision-Language Models}. In
  \bibinfo{booktitle}{\emph{Proceedings of the Computer Vision and Pattern
  Recognition Conference}}. \bibinfo{pages}{19942--19951}.
\newblock


\bibitem[Shu et~al\mbox{.}(2022)]%
        {shu2022test}
\bibfield{author}{\bibinfo{person}{Manli Shu}, \bibinfo{person}{Weili Nie},
  \bibinfo{person}{De-An Huang}, \bibinfo{person}{Zhiding Yu},
  \bibinfo{person}{Tom Goldstein}, \bibinfo{person}{Anima Anandkumar}, {and}
  \bibinfo{person}{Chaowei Xiao}.} \bibinfo{year}{2022}\natexlab{}.
\newblock \showarticletitle{Test-time prompt tuning for zero-shot
  generalization in vision-language models}.
\newblock \bibinfo{journal}{\emph{Advances in Neural Information Processing
  Systems}}  \bibinfo{volume}{35} (\bibinfo{year}{2022}),
  \bibinfo{pages}{14274--14289}.
\newblock


\bibitem[Sivakumar et~al\mbox{.}(2025)]%
        {sivakumar2025steervlm}
\bibfield{author}{\bibinfo{person}{Anushka Sivakumar}, \bibinfo{person}{Andrew
  Zhang}, \bibinfo{person}{Zaber Hakim}, {and} \bibinfo{person}{Chris Thomas}.}
  \bibinfo{year}{2025}\natexlab{}.
\newblock \showarticletitle{SteerVLM: Robust Model Control through Lightweight
  Activation Steering for Vision Language Models}. In
  \bibinfo{booktitle}{\emph{Findings of the Association for Computational
  Linguistics: EMNLP 2025}}. \bibinfo{pages}{23640--23665}.
\newblock


\bibitem[Tao et~al\mbox{.}(2025)]%
        {tao2025imgtrojan}
\bibfield{author}{\bibinfo{person}{Xijia Tao}, \bibinfo{person}{Shuai Zhong},
  \bibinfo{person}{Lei Li}, \bibinfo{person}{Qi Liu}, {and}
  \bibinfo{person}{Lingpeng Kong}.} \bibinfo{year}{2025}\natexlab{}.
\newblock \showarticletitle{Imgtrojan: Jailbreaking vision-language models with
  one image}. In \bibinfo{booktitle}{\emph{Proceedings of the 2025 Conference
  of the Nations of the Americas Chapter of the Association for Computational
  Linguistics: Human Language Technologies (Volume 1: Long Papers)}}.
  \bibinfo{pages}{7048--7063}.
\newblock


\bibitem[Tian et~al\mbox{.}(2025)]%
        {tian2025crosscheck}
\bibfield{author}{\bibinfo{person}{Baoliang Tian}, \bibinfo{person}{Yuxuan Si},
  \bibinfo{person}{Jilong Wang}, \bibinfo{person}{Lingyao Li},
  \bibinfo{person}{Zhongyuan Bao}, \bibinfo{person}{Zineng Zhou},
  \bibinfo{person}{Tao Wang}, \bibinfo{person}{Sixu Li}, \bibinfo{person}{Ziyao
  Xu}, \bibinfo{person}{Mingze Wang}, {et~al\mbox{.}}}
  \bibinfo{year}{2025}\natexlab{}.
\newblock \showarticletitle{Crosscheck-bench: Diagnosing compositional failures
  in multimodal conflict resolution}.
\newblock \bibinfo{journal}{\emph{arXiv preprint arXiv:2511.21717}}
  (\bibinfo{year}{2025}).
\newblock


\bibitem[Varma et~al\mbox{.}(2024)]%
        {varma2024ravl}
\bibfield{author}{\bibinfo{person}{Maya Varma}, \bibinfo{person}{Jean-Benoit
  Delbrouck}, \bibinfo{person}{Zhihong Chen}, \bibinfo{person}{Akshay
  Chaudhari}, {and} \bibinfo{person}{Curtis Langlotz}.}
  \bibinfo{year}{2024}\natexlab{}.
\newblock \showarticletitle{Ravl: Discovering and mitigating spurious
  correlations in fine-tuned vision-language models}.
\newblock \bibinfo{journal}{\emph{Advances in Neural Information Processing
  Systems}}  \bibinfo{volume}{37} (\bibinfo{year}{2024}),
  \bibinfo{pages}{82235--82264}.
\newblock


\bibitem[Wang et~al\mbox{.}(2025)]%
        {wang2025internvl3}
\bibfield{author}{\bibinfo{person}{Weiyun Wang}, \bibinfo{person}{Zhangwei
  Gao}, \bibinfo{person}{Lixin Gu}, \bibinfo{person}{Hengjun Pu},
  \bibinfo{person}{Long Cui}, \bibinfo{person}{Xingguang Wei},
  \bibinfo{person}{Zhaoyang Liu}, \bibinfo{person}{Linglin Jing},
  \bibinfo{person}{Shenglong Ye}, \bibinfo{person}{Jie Shao}, {et~al\mbox{.}}}
  \bibinfo{year}{2025}\natexlab{}.
\newblock \showarticletitle{Internvl3. 5: Advancing open-source multimodal
  models in versatility, reasoning, and efficiency}.
\newblock \bibinfo{journal}{\emph{arXiv preprint arXiv:2508.18265}}
  (\bibinfo{year}{2025}).
\newblock


\bibitem[Yang et~al\mbox{.}(2025)]%
        {yang2025nullu}
\bibfield{author}{\bibinfo{person}{Le Yang}, \bibinfo{person}{Ziwei Zheng},
  \bibinfo{person}{Boxu Chen}, \bibinfo{person}{Zhengyu Zhao},
  \bibinfo{person}{Chenhao Lin}, {and} \bibinfo{person}{Chao Shen}.}
  \bibinfo{year}{2025}\natexlab{}.
\newblock \showarticletitle{Nullu: Mitigating object hallucinations in large
  vision-language models via halluspace projection}. In
  \bibinfo{booktitle}{\emph{Proceedings of the Computer Vision and Pattern
  Recognition Conference}}. \bibinfo{pages}{14635--14645}.
\newblock


\bibitem[Yoon et~al\mbox{.}(2024)]%
        {yoon2024c}
\bibfield{author}{\bibinfo{person}{Hee~Suk Yoon}, \bibinfo{person}{Eunseop
  Yoon}, \bibinfo{person}{Joshua Tian~Jin Tee}, \bibinfo{person}{Mark
  Hasegawa-Johnson}, \bibinfo{person}{Yingzhen Li}, {and}
  \bibinfo{person}{Chang~D Yoo}.} \bibinfo{year}{2024}\natexlab{}.
\newblock \showarticletitle{C-tpt: Calibrated test-time prompt tuning for
  vision-language models via text feature dispersion}.
\newblock \bibinfo{journal}{\emph{arXiv preprint arXiv:2403.14119}}
  (\bibinfo{year}{2024}).
\newblock


\bibitem[Yu et~al\mbox{.}(2025)]%
        {yu2025cross}
\bibfield{author}{\bibinfo{person}{Xinmiao Yu}, \bibinfo{person}{Xiaocheng
  Feng}, \bibinfo{person}{Yun Li}, \bibinfo{person}{Minghui Liao},
  \bibinfo{person}{Ya-Qi Yu}, \bibinfo{person}{Xiachong Feng},
  \bibinfo{person}{Weihong Zhong}, \bibinfo{person}{Ruihan Chen},
  \bibinfo{person}{Mengkang Hu}, \bibinfo{person}{Jihao Wu}, {et~al\mbox{.}}}
  \bibinfo{year}{2025}\natexlab{}.
\newblock \showarticletitle{Cross-lingual text-rich visual comprehension: An
  information theory perspective}. In \bibinfo{booktitle}{\emph{Proceedings of
  the AAAI Conference on Artificial Intelligence}}, Vol.~\bibinfo{volume}{39}.
  \bibinfo{pages}{9680--9688}.
\newblock


\bibitem[Zellers et~al\mbox{.}(2019)]%
        {zellers2019recognition}
\bibfield{author}{\bibinfo{person}{Rowan Zellers}, \bibinfo{person}{Yonatan
  Bisk}, \bibinfo{person}{Ali Farhadi}, {and} \bibinfo{person}{Yejin Choi}.}
  \bibinfo{year}{2019}\natexlab{}.
\newblock \showarticletitle{From recognition to cognition: Visual commonsense
  reasoning}. In \bibinfo{booktitle}{\emph{Proceedings of the IEEE/CVF
  conference on computer vision and pattern recognition}}.
  \bibinfo{pages}{6720--6731}.
\newblock


\bibitem[Zhang et~al\mbox{.}(2025a)]%
        {zhang2025mitigating}
\bibfield{author}{\bibinfo{person}{Hao Zhang}, \bibinfo{person}{Chen Li}, {and}
  \bibinfo{person}{Basura Fernando}.} \bibinfo{year}{2025}\natexlab{a}.
\newblock \showarticletitle{Mitigating Easy Option Bias in Multiple-Choice
  Question Answering}.
\newblock \bibinfo{journal}{\emph{arXiv preprint arXiv:2508.13428}}
  (\bibinfo{year}{2025}).
\newblock

\vfill\eject

\bibitem[Zhang et~al\mbox{.}(2025b)]%
        {zhang2025robust}
\bibfield{author}{\bibinfo{person}{Zongmeng Zhang}, \bibinfo{person}{Wengang
  Zhou}, \bibinfo{person}{Jie Zhao}, {and} \bibinfo{person}{Houqiang Li}.}
  \bibinfo{year}{2025}\natexlab{b}.
\newblock \showarticletitle{Robust multimodal large language models against
  modality conflict}.
\newblock \bibinfo{journal}{\emph{arXiv preprint arXiv:2507.07151}}
  (\bibinfo{year}{2025}).
\newblock


\bibitem[Zhu et~al\mbox{.}(2025)]%
        {zhu2025ibd}
\bibfield{author}{\bibinfo{person}{Lanyun Zhu}, \bibinfo{person}{Deyi Ji},
  \bibinfo{person}{Tianrun Chen}, \bibinfo{person}{Peng Xu},
  \bibinfo{person}{Jieping Ye}, {and} \bibinfo{person}{Jun Liu}.}
  \bibinfo{year}{2025}\natexlab{}.
\newblock \showarticletitle{Ibd: Alleviating hallucinations in large
  vision-language models via image-biased decoding}. In
  \bibinfo{booktitle}{\emph{Proceedings of the Computer Vision and Pattern
  Recognition Conference}}. \bibinfo{pages}{1624--1633}.
\newblock


\end{thebibliography}
\end{document}